\documentclass{article}
\PassOptionsToPackage{numbers,compress}{natbib}
\usepackage[preprint]{neurips_2026}

\usepackage{microtype}
\usepackage{amsmath}
\usepackage{amssymb}
\usepackage{hyperref}
\usepackage{url}
\usepackage{outlines}
\usepackage{booktabs}
\usepackage{graphicx}
\usepackage[dvipsnames]{xcolor}
\usepackage[capitalize,noabbrev]{cleveref}
\usepackage{soul}
\usepackage[most]{tcolorbox}
\usepackage{fvextra}
\fvset{breaklines=true,breakanywhere=true,breaksymbolleft={\tiny\ensuremath{\hookrightarrow}}}
\usepackage{etoolbox}
\usepackage[framemethod=TikZ]{mdframed}
\usepackage{algorithm}

\definecolor{ExampleBg}{HTML}{ffffff}
\definecolor{ExampleTitle}{HTML}{629677}
\newcounter{takeaway}

\mdfdefinestyle{takeawaystyle}{
    roundcorner=5pt,
    backgroundcolor=ExampleBg,
    linecolor=ExampleTitle,
    outerlinewidth=0.3pt,
    frametitlebackgroundcolor=ExampleTitle,
    frametitlefont={\bfseries\color{white}},
    skipabove=10pt,
    frametitleaboveskip=3pt,
    frametitlebelowskip=3pt,
}

\newenvironment{takeaway}[1][]{%
    \refstepcounter{takeaway}%
    \ifstrempty{#1}%
        {\def\takeawaytitle{Takeaway~\thetakeaway}}%
        {\def\takeawaytitle{Takeaway~\thetakeaway: #1}}%
    \mdframed[style=takeawaystyle, frametitle=\takeawaytitle]%
}{%
    \endmdframed
}
\crefname{takeaway}{Takeaway}{Takeaways}

\definecolor{PaperGreen}{RGB}{85, 168, 85}

\newtcolorbox[auto counter,number within=section,crefname={box}{boxes}]{pabox}[2][]{%
title=Box 1 $\mid$,
colback=gray!10, colframe=black, sharp corners, breakable,
fontupper=\footnotesize,
subtitle style={boxrule=0.4pt,colback=cyan!50!red!25!white},title=Box ~\thetcbcounter $\mid$ #2, label={#1}}

\newtcolorbox{researchquestion}{
  breakable,
  colback=PaperGreen!20,
  colframe=PaperGreen!20,
  boxrule=0pt,
  arc=3pt,
  top=4pt, bottom=4pt,
  left=3mm, right=3mm,
  fontupper=\itshape
}

\definecolor{darkblue}{rgb}{0, 0, 0.5}
\definecolor{orange}{rgb}{0.90, 0.45, 0.00}
\hypersetup{colorlinks=true, citecolor=darkblue, linkcolor=darkblue, urlcolor=darkblue}

\title{GPU Forecasters: Language Models as Selective Surrogates for Kernel Runtime Optimization}

\author{%
  Zaid Khan$^{1}$ \quad
  Justin Chih-Yao Chen$^{1}$ \quad
  Jaemin Cho$^{2,3}$ \quad
  Elias Stengel-Eskin$^{4}$ \quad
  Mohit Bansal$^{1}$ \\
  $^{1}$UNC Chapel Hill \quad
  $^{2}$AI2 \quad
  $^{3}$Johns Hopkins University \quad
  $^{4}$University of Texas at Austin
}

\begin{document}

\maketitle

\begin{abstract}
GPU kernels are the workhorse of modern deep learning, and optimizing them
(via evolutionary search or coding agents)
usually requires repeated measurement on target hardware.
While these measurements provide the ground-truth signal necessary for kernel search, they are costly, because each evaluation of a kernel requires compilation and repeated execution on a GPU.
As improvements in LLM inference reduce the cost of writing novel kernels and LLM-driven searches scale to large search budgets,
on-device evaluation becomes a bottleneck.
To address this, we study how LLMs can serve as selective GPU surrogates for kernel evaluation, by forecasting the performance of proposed kernels.
A useful surrogate should be accurate, and it should be selective, by knowing when it could be wrong, and deferring to the GPU. 
To evaluate surrogates, we measure whether their forecasts are accurate, calibrated, and practically useful for recovering fast kernels under limited GPU-measurement budgets.
Next, we study whether reinforcement learning can improve forecast accuracy and confidence calibration. 
Our experiments demonstrate that LLMs can accurately forecast relative kernel performance, that their utility can be improved through reinforcement learning.
Used inside a kernel search, the surrogate lets the search consider several times as many candidates 
under the same GPU evaluation budget, and that leads to finding faster kernels than an equal-budget baseline.
These results suggest that LLMs can play a broader role in kernel optimization, by acting as virtual models of a GPU rather than solely as kernel generators for search.\footnote{Code: \href{https://github.com/codezakh/gpu-forecasters}{github.com/codezakh/gpu-forecasters}}
\end{abstract}

\section{Introduction}
GPU kernels are the low-level programs that run tensor operations on GPUs. They are critical to modern deep learning, and small differences in their efficiency translate into large differences in the cost and feasibility of training and inference.
Writing high-performance kernels is among the most specialized tasks in software engineering, requiring detailed knowledge of memory hierarchies, tensor core instructions, and architecture-specific tradeoffs that vary across hardware generations.
Recent work has applied LLMs to this problem, with LLM-powered search methods such as CUDA Agent~\citep{cuda_agent}, TTT-Discover~\citep{ttt-discover}, and K-Search~\citep{k_search} demonstrating strong results.
      The core of these search procedures is a generation-evaluation cycle where generated kernels are evaluated for correctness and speed.
However, each evaluation
      is expensive, requiring compilation and on-GPU execution, searches use many evaluations (e.g. \citep{ttt-discover} use 25,600 H100 GPU evaluations for optimizing the \texttt{TriMul} kernel).
In addition, evaluation throughput can be an order of magnitude slower ($\frac{1}{10}$-th) than generation throughput.
These factors make on-GPU evaluation a bottleneck for kernel optimization.
Although a large body of prior work has focused on proposing new kernels \citep{agentic_variation_operators,cuda_agent,k_search,kernelbench,cutegen, cutlass_sol},
the problem of reducing their evaluation cost has received less attention. 
One way to reduce number of on-device evaluations is to use a model of kernel execution to reason about what the outcome of running a kernel would be.
World models that allow an agent to compare and choose possible actions before executing them in the environment have been useful in other domains ~\citep{sutton_dyna, world_models, dreamerv3}. 
Recent work has studied whether LLMs can learn world models from small interaction budgets in game and puzzle environments~\citep{autumnbench, poe_world, onelife, arc_agi_2}, but has not investigated physical hardware or kernel optimization.
Meanwhile, LLMs have shown impressive ability to reason about how general purpose programs will execute~\citep{cruxeval, cwm_meta, prasadlearning}.
Because contemporary LLMs can already guide kernel search, we ask to what extent these code-reasoning capabilities can be adapted to predict how a physical device will respond to candidate kernels.
We refer to such a model as a \textit{surrogate} of the device, following its use in surrogate-assisted optimization~\citep{jones_ego}.
Here the model's role is to forecast the device's response to a candidate kernel, rather than to generate the next candidate kernel (which previous work focuses on).
However, in practice, there is a trade-off between forecasting and real on-device execution: on cases where the model's forecast is poor, there may be no substitute for real execution.  
This connects to \emph{selective prediction} \citep{chow1957optimum, elyaniv_wiener}, where a model is evaluated on the subset of inputs it keeps, and to active learning \citep{settles2009active}, where labels are acquired selectively under a budget.
This framing raises a key research question:

\begin{researchquestion}
\centering
Can we use LLMs as selective surrogates of a physical GPU that make accurate forecasts of kernel execution and indicate when those forecasts can be trusted?
\end{researchquestion}
\begin{figure}[t]
  \centering
  \includegraphics[width=\linewidth]{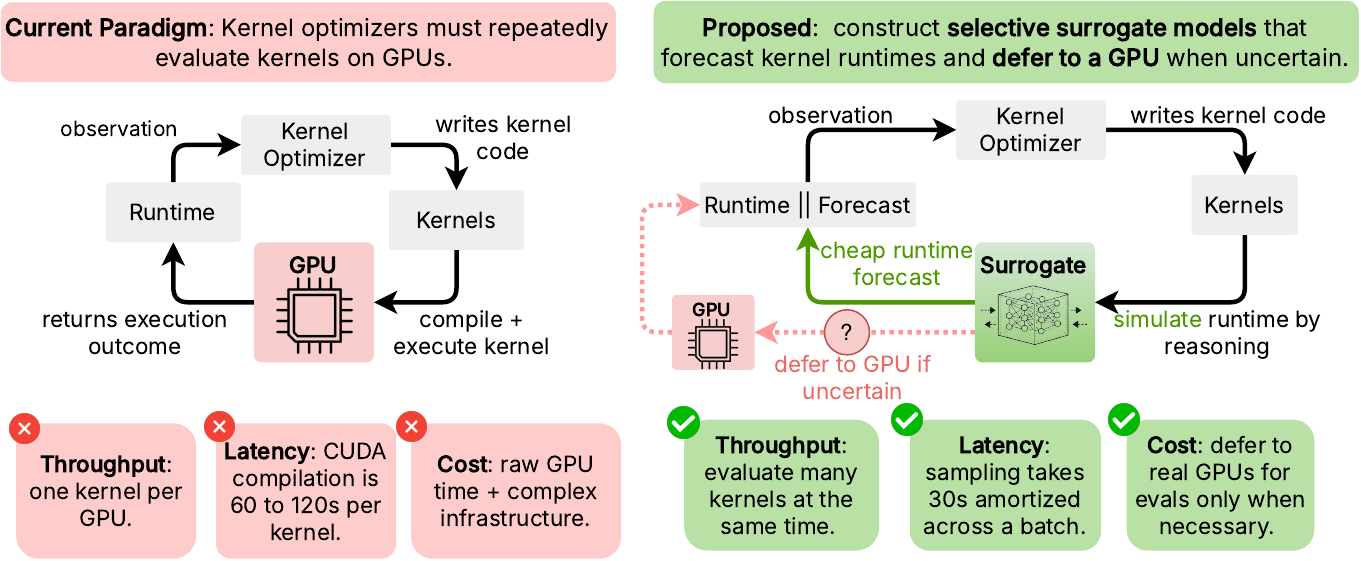}
  \caption{
  Left: LLM-driven kernel optimizers use physical GPU measurements as feedback for candidate kernels, making search costly when many candidates must be compiled, executed, and profiled.
  Right: a learned surrogate forecasts execution outcomes for candidate kernels and flags uncertain cases for real GPU measurement.
  This motivates using the surrogate where its forecasts are reliable and reserving device measurement for candidates where the forecast is uncertain.
  }
  \label{fig:teaser}
\end{figure}

With a view to addressing this question, we evaluate a range of LLMs as calibrated GPU surrogates.
Forecasting the device's exact response to a candidate kernel would require simulating its microarchitectural execution, which is not tractable.
We instead formulate the forecast target as the relative change in the device's response between a reference kernel and another kernel from the same optimization problem.
We evaluate these forecasts through four questions tied to kernel search: (1) can the surrogate prioritize promising candidates for scarce GPU measurements, (2) can its confidence identify forecasts that are safe to trust, (3) can reinforcement learning improve these properties without weakening the ranking signal, and (4) does using the surrogate inside a kernel search find faster kernels at a fixed GPU-measurement budget?
Together, these questions characterize both the forecast quality of the surrogate and the measurement-budget tradeoff it would expose to a kernel optimizer.
Finally, we apply reinforcement learning to an open-weights reasoning model, using rewards for calibration and correctness, and study how training shifts its ranking, confidence reliability, and forecast quality.
We find that LLMs can forecast useful GPU outcomes, RL teaches the surrogate where to spread uncertainty, and calibration training improves reliability as well as ranking quality.
A practical consequence of this framing is that the training data for a surrogate is a by-product of running search itself: every measured candidate already carries the (reference, candidate, hardware, speedup) tuple a surrogate consumes, so a long-running kernel search yields its own training set without separate data collection.

This paper makes five contributions.
(1) We show that under the right prediction objective, \textbf{LLMs achieve non-trivial accuracy in forecasting how a GPU responds to a candidate kernel}.
(2) We cast \textbf{GPU forecasting as a selective prediction} problem, in which the surrogate must report calibrated uncertainty so that downstream search can decide which predictions are reliable enough to act on.
(3) We apply \textbf{reinforcement learning to an open-weights surrogate} and characterize how training shifts its ranking, calibration, and forecast quality.
(4) We show that the surrogate \textbf{improves end-to-end kernel search}, finding faster kernels than an equal-GPU-budget baseline on most tasks by letting the search consider more candidates per measurement.
(5) We build a reusable, multi-purpose collection of \textbf{12,388 LLM-generated GPU kernels with measured runtimes}, spanning 118 problems, CUDA and Triton, three GPU types, and four generation methods, costing 400M LLM tokens and 600 GPU-hours. The full release supports broader work on GPU performance
prediction, kernel search, and LLM-generated low-level code.
More broadly, GPU surrogates connect kernel optimization to efforts to build neural simulators of complex physical and virtual systems~\citep{neural_os,neural_computers}.
The long-term promise of such models is not only faster prediction, but the ability to generate training data and support model-based optimization without querying the original system at every step.
A surrogate offers higher throughput and lower per-query cost than a GPU, in exchange for forecast noise that calibrated uncertainty estimates can help manage.

\begin{figure}
    \centering
    \includegraphics[width=1.0\linewidth]{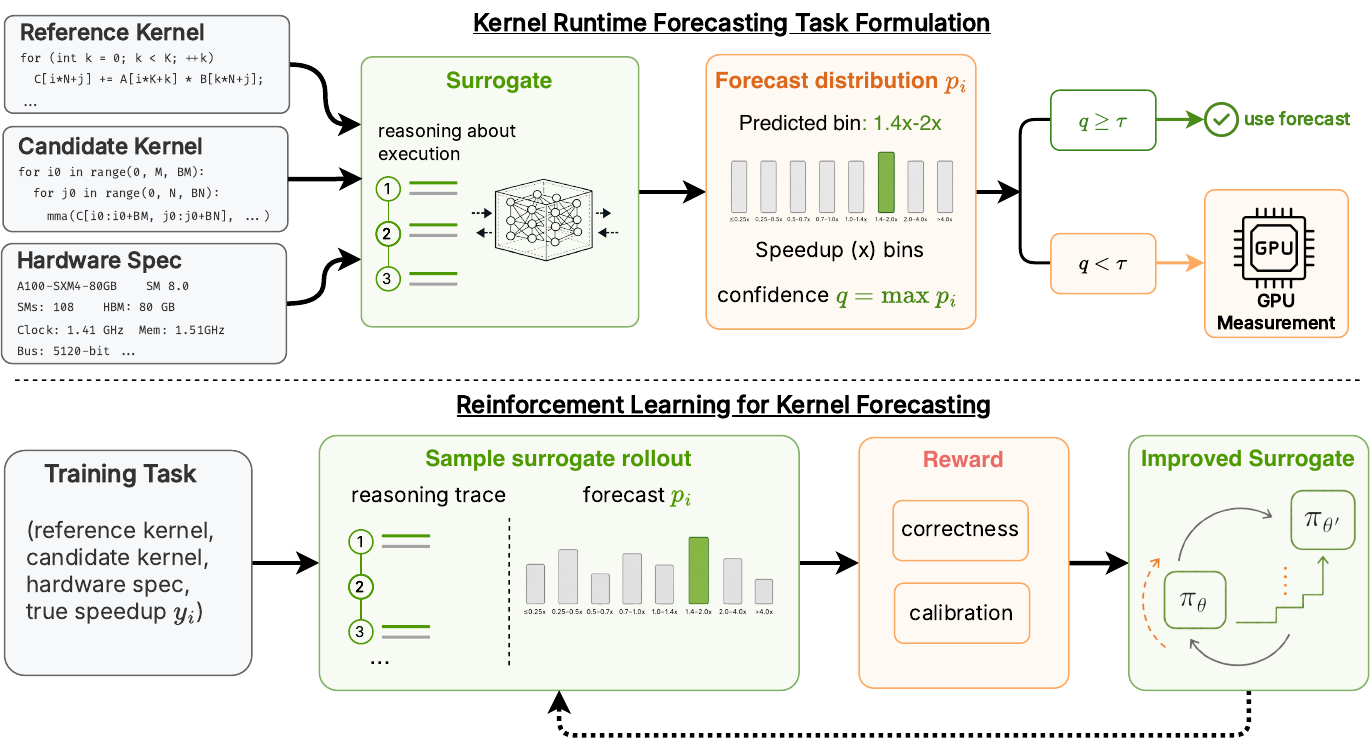}
    \caption{Our method casts kernel evaluation as selective forecasting over discrete speedup bins.
    Top: the surrogate produces a forecast distribution over speedup bins and defers low-confidence candidates to a GPU measurement.
    Bottom: we train the surrogate with reinforcement learning, with rewards for correctness and calibration of the forecast distribution.
    We use this setup to study how reinforcement learning changes the surrogate's behavior.}
    \label{fig:method}
\end{figure}
\section{Method: LLM as Virtual GPU Surrogate}
\subsection{Task Formulation}
\label{sec:task-formulation}
\paragraph{Kernel Evaluation.}
A kernel optimization problem starts with a target operation and a reference implementation that defines the correct outputs on a fixed set of test inputs.
For example, one task in our evaluation is \texttt{TriMul},\footnote{\url{https://www.gpumode.com/leaderboard/496}} a triangle multiplicative update forward pass from AlphaFold-style protein structure models~\citep{alphafold3}.
LLM-driven kernel search procedures can rapidly generate thousands of candidate kernels intended to implement the same operation, as shown in \cref{fig:teaser}.
Ordinarily, each candidate kernel is compiled, run on the test inputs, checked against the reference outputs, and timed on the target GPU.
This evaluation is expensive because it includes compilation, correctness checking, and on-device profiling rather than only kernel execution.
For CUDA kernels, compilation can take tens of seconds per candidate, so a search that evaluates thousands of candidate kernels spends substantial wall-clock time (\cref{app:throughput}) before accounting for GPU execution and infrastructure overhead.
The GPU outcome for a candidate kernel has two parts: validity and speed. 
Invalidity means that the candidate fails to compile, fails at runtime, or fails correctness checking.
If the candidate is valid, we measure its runtime and compare it to the reference: speedup above $1\times$ means the candidate is both correct and faster, while speedup below $1\times$ means it is correct but slower.

\paragraph{Forecasting Relative Performance}
For candidate $i$, let $x_i=(r,k_i,h)$ denote the (reference kernel implementation, candidate kernel implementation, and hardware specification).
Let $S_i=T_{\mathrm{ref}}/T_i$ be its measured speedup over the reference.
If the candidate is valid, $S_i>1$ means the candidate is faster and $S_i<1$ means it is slower.
We implement the \textit{surrogate} as an LLM prompted to take $x_i$ as input, generate a brief chain of reasoning, and forecast the candidate's speedup bin on the device.

Predicting an absolute runtime in seconds from source code alone is intractable, as it would require modeling compiler behavior, launch overheads, memory hierarchy effects, scheduling, resource occupancy, and other hardware-dependent execution details.
We instead define the prediction target as the candidate's speedup $S_i$ relative to the reference, discretized into eight ordinal bins.
We use logarithmic (half-octave) speedup bins rather than linearly spaced bins because the surrogate is forecasting from source code rather than executing the kernel.
Fine-grained numeric speedups, such as distinguishing a $2\times$ improvement from a $3\times$ improvement, are difficult to infer reliably without running the kernel.
Coarse factor-level judgments are more realistic: the model can estimate whether a candidate is much slower, near baseline, moderately faster, or much faster.
These intervals make the forecast target meaningful and tractable while preserving the distinctions a kernel optimizer needs; the full bin table is provided in \cref{tab:speedup-bins}.
The surrogate emits a probability distribution over the eight bins:
\[
  p_i=(p_{i,1},\ldots,p_{i,8}), \qquad p_{i,b} \geq 0, \qquad \sum_{b=1}^{8}p_{i,b}=1.
\]
The predicted bin is $\hat{y}_i=\arg\max_b p_{i,b}$, and the stated confidence is $q_i=\max_b p_{i,b}$.
In the structured tool call, this forecast consists of the eight probabilities $p_i$, the predicted bin $\hat{y}_i$, and a short rationale.

\paragraph{Selective Use.}
We use the term \textit{selective surrogate} for a surrogate whose forecasts can be used selectively rather than accepted on every candidate.
This follows the selective-prediction setting \citep{chow1957optimum, elyaniv_wiener}, where a model abstains on uncertain inputs.
In kernel search, the analogous use would be to trust the surrogate on candidates where its forecast is reliable and reserve GPU measurements for cases where the surrogate is uncertain.

We therefore evaluate whether the surrogate's stated confidence is informative enough to support threshold-based use (detailed in \cref{sec:confidence}): the surrogate emits the same distributional forecast as above, and a kernel optimizer could accept forecasts with $q_i$ above a confidence threshold while measuring the remaining candidates on the device.
Because the selective signal comes from the forecast distribution itself, improving the quality of that distribution is the training problem we study next.

\subsection{Reinforcement Learning to Improve Forecasting}
\label{sec:rl-training}
We use reinforcement learning (RL) to study how training changes an open-weights surrogate, GPT-OSS-20B~\citep{gptoss}.
This forecasting task is a natural fit for reinforcement learning with verifiable rewards.
We ask the model to reason about a candidate kernel and predict its runtime bin; after the rollout, we can compare that prediction to the measured runtime bin from the GPU.
We then reinforce reasoning chains that lead to better forecasts, using rewards defined on the final structured prediction.

The training examples have the same form as the evaluation queries: a reference kernel, a candidate kernel, the target hardware, and the candidate's measured speedup bin relative to the reference.
We do not train on failures; every RL example has a successful device measurement and a label in one of the eight speedup bins.
Let $y_i \in \{1,\ldots,8\}$ be the measured speedup bin for training example $i$.

A rollout is one sampled chain of reasoning that ends in a structured forecast tool call.
Let $p_i$ be the probability distribution returned by the forecast schema defined in \cref{sec:task-formulation}, and let $\hat{y}_i$ be its predicted bin.
If the rollout does not produce a parseable forecast, it receives zero reward.

\paragraph{Correctness Reward.}
This reward directly trains the surrogate to maximize the probability of the true measured speedup bin:
$r_{\mathrm{corr}} = \mathbf{1}[\hat{y}_i=y_i].$
Exact-bin correctness ignores the rest of the probability distribution.
Following the general idea of using RL rewards to train language models to report useful uncertainty~\citep{stengel2024lacie, rlcr} we add two distributional rewards to shape the full distribution while keeping exact-bin correctness as the primary task signal.

\paragraph{Categorical Calibration (Brier)}
The Brier score treats the eight bins as distinct classes.
It compares the predicted probabilities to a one-hot target, so it rewards probability on the measured bin and penalizes probability elsewhere:
\[
  \mathrm{Brier}(p_i,y_i)
  = \frac{1}{2}\sum_{b=1}^{8}\left(p_{i,b} - \mathbf{1}[b=y_i]\right)^2,
  \qquad
  r_{\mathrm{corr+Brier}} = r_{\mathrm{corr}} + 1 - \mathrm{Brier}(p_i,y_i).
\]

\paragraph{Distance-Aware Calibration (CRPS)}
Alternatively, the continuous ranked probability score (CRPS) uses the ordering of the speedup bins.
Let $F_k(p_i)=\sum_{b=1}^{k}p_{i,b}$ be the cumulative predicted probability up to bin $k$.
CRPS compares this cumulative distribution to the cumulative one-hot target, so mass one bin away from the truth is penalized less than mass many bins away:
      \[
        \mathrm{CRPS}(p_i,y_i)
        = \frac{1}{7}\sum_{k=1}^{7}\left(F_k(p_i) - \mathbf{1}[y_i \leq k]\right)^2,
        \qquad
        r_{\mathrm{corr+CRPS}} = r_{\mathrm{corr}} + 1 - \mathrm{CRPS}(p_i,y_i).
      \]
Implementation details for the RL runs, including model, training settings, rollout count, decoding settings, and dataset construction, are given in \cref{app:rl-implementation}.

\section{Experiments}
\label{sec:experiments}

\subsection{Metrics}
\label{sec:metrics}

We evaluate each surrogate with three metrics: whether it ranks fast kernels highly, whether its confidence is calibrated, and how large its speedup mistakes are.

\textbf{Speedup recovered.} This metric asks how close a search would get to the best available kernel if it were to only measure the top $k$ candidates that the surrogate ranks highest.
For candidate $i$, the surrogate's probability distribution over speedup bins defines an expected speedup score $\hat{S}_i$.
For a measurement budget $c$, we take the top $c\%$ of candidates under this score and report the best measured speedup in that subset as a percentage of the best measured speedup available in the full candidate set.
This can also be thought of as a measure of regret: when ranking kernels and selecting the top $c\%$, how much worse is the best kernel in that top $c\%$ than the best kernel present in the superset of candidates. 
A perfect ranking would recover $100\%$ of the speedup at every budget. 
We average this value over budgets $\mathcal{C}=\{1\%,5\%,10\%,25\%,50\%\}$.
\[
  \hat{S}_i = \sum_{b=1}^{8} p_{i,b}\,m_b,
  \qquad
  \mathrm{Recovered} = \frac{100}{|\mathcal{C}|}\sum_{c \in \mathcal{C}}\frac{\max_{i \in \mathrm{Top}_c(\hat{S})} S_i}{\max_i S_i}.
\]
Here $m_b$ is the representative speedup assigned to bin $b$, defined in \cref{tab:speedup-bins}.
A surrogate scores well when the candidates it ranks near the top include the fastest kernels found by exhaustive measurement.

\textbf{Expected Calibration Error (ECE).} This metric asks whether the surrogate's reported confidence matches its empirical accuracy.
For each candidate, confidence is $q_i=\max_b p_{i,b}$, the probability assigned to the speedup bin the surrogate predicts.
We sort predictions into ten confidence buckets following \citet{naeini2015obtaining}; within each bucket, $\mathrm{conf}(B_m)$ is the average confidence and $\mathrm{acc}(B_m)$ is the fraction of predictions whose speedup bin exactly matches the measured bin.
\[
  \mathrm{ECE} = \sum_{m=1}^{10}\frac{|B_m|}{n}\left|\mathrm{acc}(B_m) - \mathrm{conf}(B_m)\right|.
\]
A surrogate that states $70$ percent confidence on a set of predictions should be right on about $70$ percent of them; lower ECE means this agreement is closer.

\textbf{Forecast error.} This metric asks how numerically far the predicted speedup is from the measured speedup.
We convert the predicted speedup bin $\hat{y}_i$ to its representative speedup $m_{\hat{y}_i}$ and compute the mean absolute gap from the measured speedup $S_i$ in $\times$ units.
\[
  \mathrm{ForecastError} = \frac{1}{n}\sum_{i=1}^{n}\left|m_{\hat{y}_i} - S_i\right|.
\]

\subsection{Setup}
\label{sec:setup}
We evaluate surrogate forecasts on a set of 480 Triton kernels covering six tasks from the GPU Mode\footnote{\url{https://www.gpumode.com/home}} kernel optimization competition: TriMul, cross-entropy, two Gated DeltaNet forward kernels, GDN-W/U recompute, and FP8 quantization.
We target these tasks because they are realistic workloads with human-engineered reference implementations that humans have demonstrably sped up, providing an attainable optimization target, and because prior LLM-driven kernel search work~\citep{ttt-discover, k_search} has used the same competition as its evaluation surface.
Each candidate is compared against its task's single competition reference and the runtime distribution is approximately uniform across all tasks so that no single performance regime dominates the evaluation.
All device measurements use NVIDIA A100 GPUs, and we sample three times from each surrogate at temperature $1.0$, with tables reporting mean $\pm$ population standard deviation across the three repeats.
For RL training we draw on a separate pool of successful candidates generated by GPT-OSS-20B max-reward tree search runs over the same six tasks, disjoint from the held-out set.
Full eval-set construction, training-pool construction, and the compute footprint of the underlying searches are deferred to \cref{app:eval-set,app:rl-implementation}.

\subsection{How do LLMs perform on forecasting of GPU outcomes?}
\label{sec:forecast-shape}
\begin{table}[t]
  \centering
  \caption{\textbf{Off-the-shelf LLM surrogates recover most of the available speedup under small GPU-measurement budgets, while calibration varies substantially.}
  Speedup recovered is averaged over top-ranked $1\%$--$50\%$ GPU-measurement budgets.
  ECE and forecast error measure calibration and prediction error; lower is better.
  Cells show mean $\pm$ standard deviation across three repeats.}
  \label{tab:headline}
  \small
  \setlength{\tabcolsep}{4pt}
  \begin{tabular}{lccc}
    \toprule
    Surrogate & Speedup recovered (\%) $\uparrow$ & ECE $\downarrow$ & Forecast error $\downarrow$ \\
    \midrule
    Gemini-3 Flash & $93.12 \pm 0.97$ & $0.624 \pm 0.002$ & $0.421 \pm 0.021$ \\
    GPT-OSS-120B & $86.85 \pm 2.51$ & $0.479 \pm 0.005$ & $0.632 \pm 0.029$ \\
    GPT-OSS-20B & $82.88 \pm 3.85$ & $0.441 \pm 0.011$ & $0.643 \pm 0.011$ \\
    DeepSeek-V4~\citep{deepseekv4} & $79.86 \pm 3.98$ & $0.401 \pm 0.009$ & $0.805 \pm 0.023$ \\
    \bottomrule
  \end{tabular}
\end{table}
\Cref{tab:headline} reports speedup recovered, the fraction of the best achievable speedup a search finds when it GPU-tests only the surrogate's top-$c\%$ of candidates, averaged across budgets from $1\%$ to $50\%$.
The off-the-shelf GPT-OSS-20B scores $82.9\%$, which corresponds to finding candidates within $30\%$ of the true best at a $1\%$ measurement budget and within $6\%$ of it at $50\%$.

\begin{takeaway}[LLMs can forecast useful GPU outcomes]
Ranking candidates by an LLM's forecast recovers a substantial amount of the available speedup while running only a small fraction of candidates on the GPU ($1\%$--$50\%$).
\end{takeaway}

\subsection{How does RL affect the surrogate?}
\label{sec:rl-surrogate}

\begin{figure}
    \centering
    \includegraphics[width=1.0\linewidth]{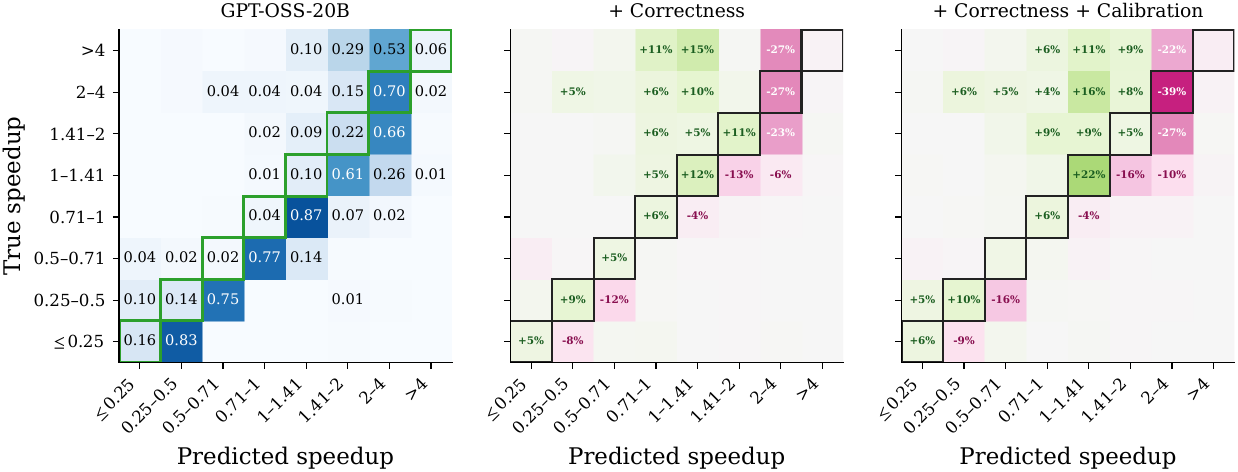}
    \caption{\textbf{The base surrogate overestimates slow kernels and underestimates the fastest kernels, while RL changes where those errors go.}
    Rows are measured speedup bins and columns are predicted bins for GPT-OSS-20B; each cell is the fraction of examples in that measured bin receiving that prediction.
    The middle and right panels show how training changes those row fractions relative to the base (\% change): green cells gain probability mass after training and magenta cells lose it.}
    \label{fig:confusion-matrix}
\end{figure}

\paragraph{What does RL training change about the surrogate, and at what cost?}
\label{sec:rl-mechanism}
\Cref{fig:confusion-matrix} shows that the untrained GPT-OSS 20B predicts slow / moderate kernels will be faster than they are, while predicting $>4\times$ kernels to be slower than they are. 
Kernels $4\times$ faster than the reference are likely hard to spot for the model.
RL redistributes probability mass across speedup bins, and \cref{tab:rl-deltas} shows the tradeoff.

\Cref{tab:rl-deltas} compares each RL reward to the GPT-OSS-20B base.
All three rewards improve calibration, but they differ in search utility: correctness-only training loses $1.29$\% in speedup recovered, CRPS is roughly neutral, and Brier gains $2.06$\%.
The correctness reward only checks whether the argmax bin is correct, so it can improve exact-bin decisions while ignoring how useful the rest of the probability vector is.
Brier and CRPS instead score the full distribution $p_i$.
That matters for search because candidates are ranked by expected speedup, $\hat{S}_i=\sum_b p_{i,b}m_b$, not just by the argmax bin.
A probability vector with informative mass across nearby bins can separate candidates that share the same predicted bin while a binary vector cannot.
The tradeoff is that all trained variants increase forecast error.
Forecast error uses only the argmax-bin representative $m_{\hat{y}_i}$, so it can worsen when training shifts the most likely bin even if the full distribution becomes better calibrated or more useful for ranking.
Calibration is important here, as it affects decision-making about when to defer, which kernels to prioritize, and other actions which depend on the quality of the forecast distribution.

\begin{table}[t]
  \centering
  \caption{\textbf{RL changes several surrogate properties at once.}
  All entries are changes from the untrained GPT-OSS-20B base; rows differ only in reward.
  Length is reasoning chain length. Latency is per-call latency.
  }
  \label{tab:rl-deltas}
  \small
  \setlength{\tabcolsep}{3pt}
  \resizebox{\linewidth}{!}{%
  \begin{tabular}{ccccccc}
    \toprule
    \multicolumn{2}{c}{Rewards} & \multicolumn{5}{c}{Change vs. GPT-OSS-20B base} \\
    \cmidrule(lr){1-2}\cmidrule(lr){3-7}
    Correctness & Calibration & Speedup recovered (\%) $\uparrow$ & ECE $\downarrow$ & Forecast error $\downarrow$ & Length (tokens) $\downarrow$ & Latency (s) $\downarrow$ \\
    \midrule
    \checkmark & -- & $-1.29$ & $-0.031$ & $+0.118$ & $-812$ & $-3.39$ \\
    \checkmark & CRPS & $-1.05$ & $-0.025$ & $+0.100$ & $+184$ & $+17.80$ \\
    \checkmark & Brier & $+2.06$ & $-0.106$ & $+0.130$ & $-982$ & $-5.02$ \\
    \bottomrule
  \end{tabular}}
\end{table}

\begin{takeaway}[RL teaches the surrogate where to spread uncertainty]
The base surrogate overpredicts slow kernels and underpredicts the fastest kernels.
RL spreads uncertainty across plausible speedup bins, improving confidence reliability but increasing forecast error.
Reward choice sets the tradeoff between search utility and inference cost.
\end{takeaway}

\begin{figure}
    \centering
    \includegraphics[width=1.0\linewidth]{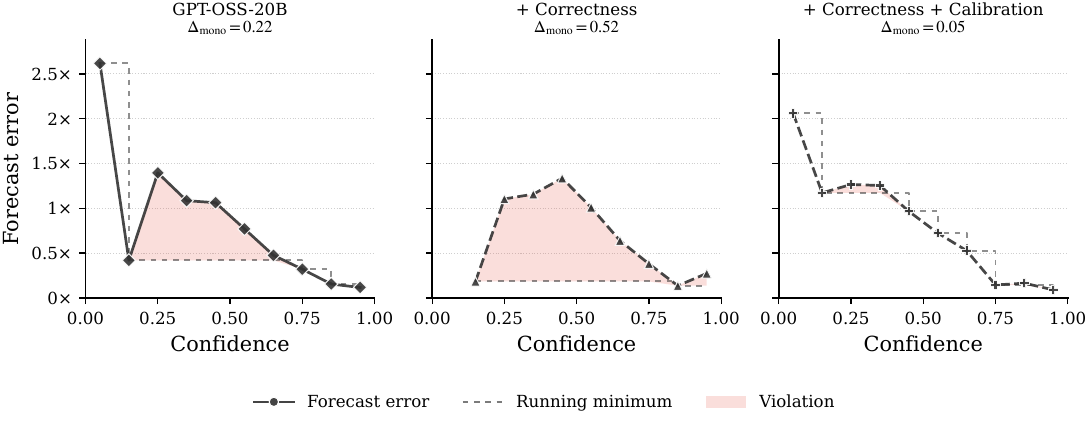}
    \vspace{-1.75em}
    \caption{\textbf{Calibration rewards make confidence track forecast error, while correctness-only RL weakens this relationship.}
    Examples are bucketed by stated confidence, $q_i=\max_b p_{i,b}$.
    The $y$-axis is mean forecast error within each bucket; lower and more decreasing curves indicate more reliable confidence.
    Shading marks violations of monotone decrease, summarized by $\Delta_{\mathrm{mono}}$ with $0$ indicating no violation. 
    }
    \label{fig:reliability}
\end{figure}

\paragraph{Does stated confidence track forecast quality?}
\label{sec:confidence}
The surrogate's confidence is useful for selective use only if it correlates with forecast quality, that is, if a candidate the surrogate labels confidently is one it really gets right.
\Cref{fig:reliability} tests this by plotting mean forecast error against stated confidence, bucket by bucket.
A desirable surrogate would produce a curve that decreases monotonically (i.e., forecast error is monotonically lower for higher-confidence predictions), and $\Delta_{\mathrm{mono}}$ quantifies this by measuring how far the curve deviates from monotonicity, with $0$ meaning monotone-decreasing and $1$ meaning monotone-increasing; lower is better. 
The off-the-shelf base satisfies the property roughly ($\Delta_{\mathrm{mono}} = 0.22$), with high-confidence forecasts off by less than $0.5\times$ on average and low-confidence forecasts off by more than $2\times$.
Training on correctness alone destroys it ($\Delta_{\mathrm{mono}} = 0.52$): the model becomes confident on candidates it gets wrong by a wide margin.
Adding a Brier shaping term to the same correctness reward improves it even beyond the base model ($\Delta_{\mathrm{mono}} = 0.05$).

\begin{takeaway}[Calibration training improves reliability]
Correctness-only RL makes confidence less reliable: higher confidence no longer consistently means lower forecast error.
Adding a Brier calibration term reverses this effect, making forecast error nearly monotone in stated confidence.
\end{takeaway}

\begin{figure}
    \centering
    \includegraphics[width=1.0\linewidth]{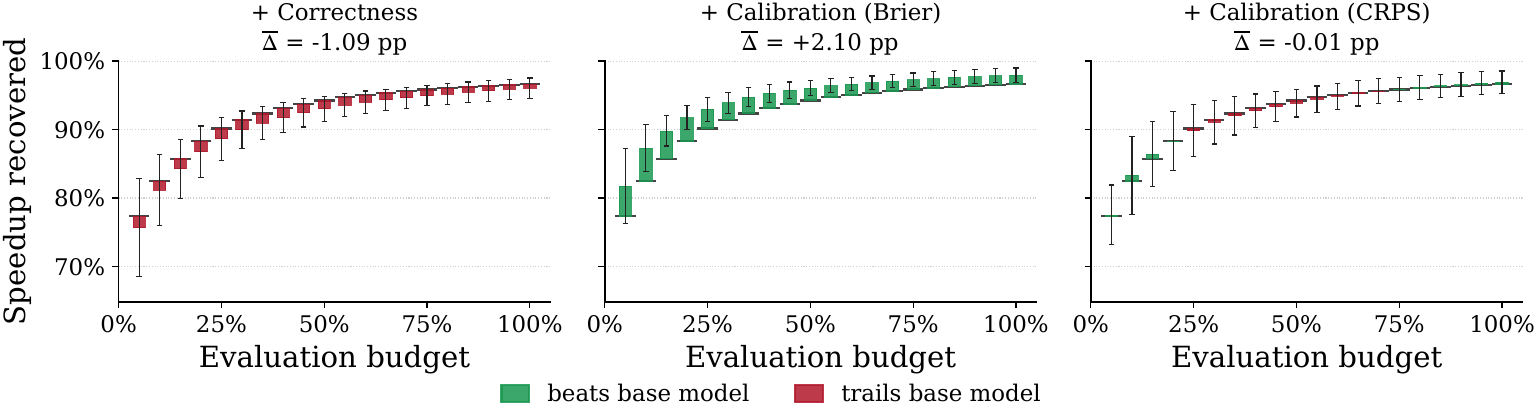}
    \caption{\textbf{Brier calibration improves the speedup found under most measurement budgets; correctness-only and CRPS training do not.}
    Budgets are the fraction of candidates GPU-measured after sorting by predicted expected speedup.
    Gray dashes show the GPT-OSS-20B base; colored bars show trained-minus-base speedup recovered, with green above and red below the base.
    Error bars show $\pm 1\sigma$ across repeats, and $\overline{\Delta}$ is the area-averaged change across the plotted budget curve in percentage points.}
    \label{fig:recovered-delta}
\end{figure}
\paragraph{Under a fixed measurement budget, how much of the true best speedup does ranking by the surrogate recover?}
\label{sec:recovery}
\Cref{fig:recovered-delta} asks how useful the surrogate would be inside a kernel search loop where GPU measurements are the scarce resource.
A searcher would rank generated candidates by predicted expected speedup, measure from the top of that list downward, and hope to find a fast kernel before exhausting its budget.
We therefore simulate this process at different measurement budgets and report the fraction of the best available speedup recovered.
\Cref{tab:headline} summarizes this metric over the small-budget points $1\%, 5\%, 10\%, 25\%, 50\%$, while \Cref{fig:recovered-delta} shows the trained-minus-base change across the full plotted sweep and reports its area-averaged delta.
The base is already strong, recovering $77\%$ of the best available speedup after measuring the top-ranked $5\%$ of candidates and $94\%$ after measuring $50\%$.
Across the full plotted curve, Brier-shaped training averages $+2.10$ pp, CRPS shaping averages $-0.01$ pp, and correctness-only averages $-1.09$ pp.
The pattern within proper scoring rules is unexpected.
CRPS gives partial credit when the prediction is close to the right bin, while Brier penalizes every wrong bin equally, so a priori one would expect CRPS to give the better ranking.
The data shows the opposite: bin-order awareness in the loss does not, on its own, lift the candidate ranking the surrogate produces.

\begin{takeaway}[Calibration training improves ranking quality]
Brier-shaped training improves more than reported confidence: it also improves recovered speedup under measurement budgets.
The gain shows that calibration training changes the forecast distribution in a way that strengthens the ranking signal used by budgeted kernel search.
\end{takeaway}

\subsection{Does the surrogate improve a real kernel search?}
\label{sec:extrinsic}
\begin{figure}
    \centering
    \includegraphics[width=1.0\linewidth]{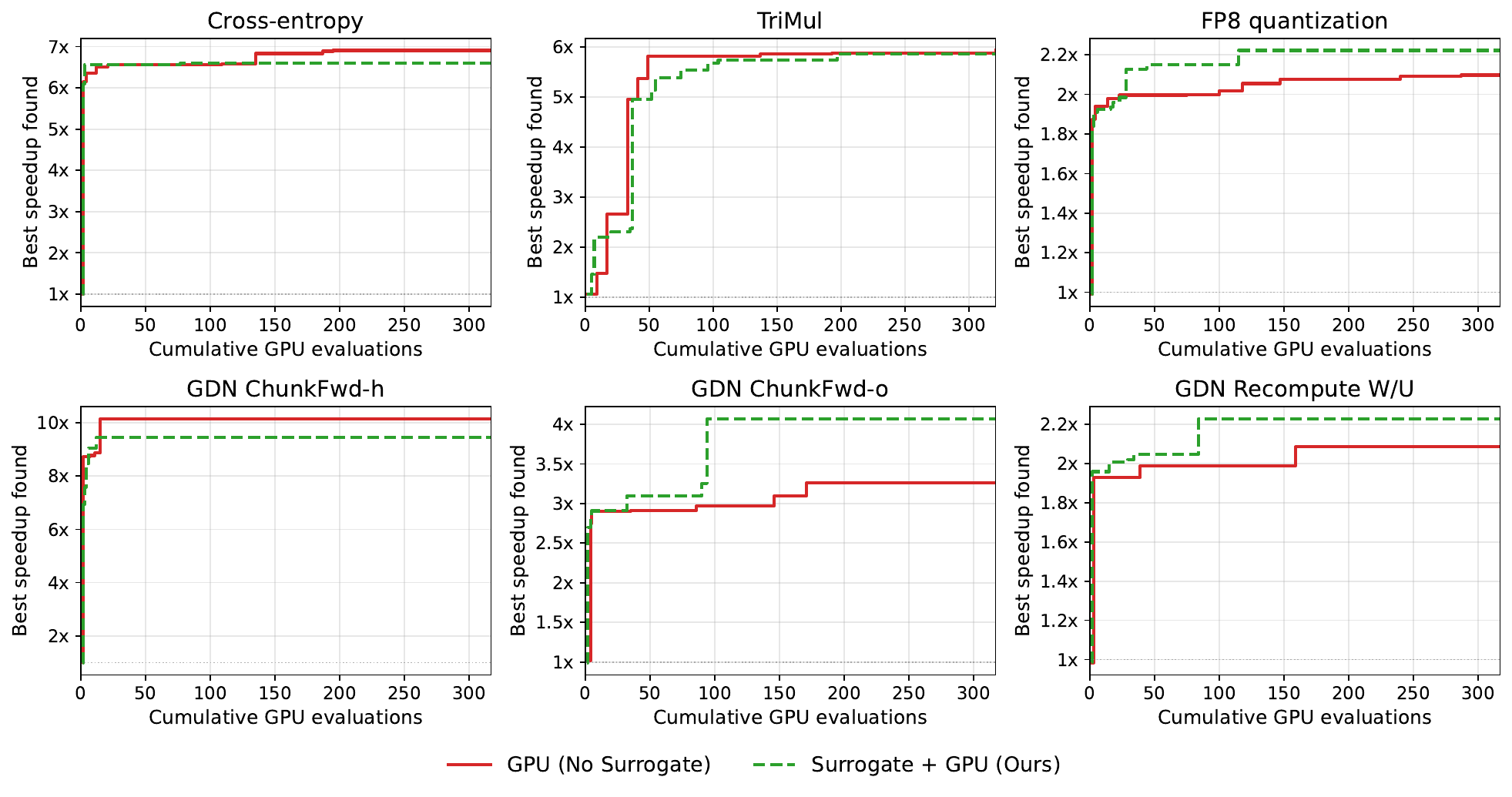}
    \caption{\textbf{The search with the surrogate matches or improves on the baseline's best speedup on four of the six tasks, often at substantially fewer GPU measurements, with gaps of $5\%$ and $7\%$ on the remaining two.}
    Each panel is one of the six tasks of \cref{sec:setup}.
    The $y$-axis is the best measured kernel speedup so far.
    The $x$-axis is the cumulative number of GPU measurements consumed by the search.
    Without the surrogate, the search measures every candidate the mutator proposes.
    With the surrogate, the search proposes four times as many candidates per parent and measures only the quarter ranked highest by the Brier-trained GPT-OSS-20B surrogate.
    Both procedures spend the same number of GPU measurements per step.
    }
    \label{fig:extrinsic-curves}
\end{figure}

We compare two PUCT searches (\cref{app:puct}) on the six tasks of \cref{sec:setup}.
Without the surrogate, the search measures every candidate the mutator proposes.
With the surrogate, the search proposes four times as many candidates per parent and measures only the quarter ranked highest by the surrogate, where the ranking uses the expected-speedup score of \cref{sec:metrics}.
Both procedures spend the same number of GPU measurements per step.
We use Gemini-3 Flash as the mutator and the Brier-trained GPT-OSS-20B from \cref{sec:rl-surrogate} as the surrogate.

\Cref{fig:extrinsic-curves} plots the best measured speedup against cumulative GPU measurements for each procedure on each task.
On three of the six tasks (FP8 quantization, GDN ChunkFwd-o, GDN Recompute W/U), the search with the surrogate finds a faster kernel than the baseline and reaches or exceeds the baseline's best speedup using substantially fewer measurements.
On TriMul the two procedures reach essentially the same speedup.
On the remaining two tasks the baseline finds a faster kernel. On cross-entropy the gap is about $5\%$, and on GDN ChunkFwd-h both procedures saturate within the first 15 measurements with the baseline $7\%$ above.
The two searches where the baselines win are cases where the search appears saturated, as kernels found in the first steps of search are already very close to the final best kernel.

\begin{takeaway}[A surrogate lets a search explore with fewer GPU evaluations]
The surrogate allows search to explore more candidates than the GPU evaluation budget allows while prioritizing GPU evaluations for promising candidates.
\end{takeaway}

\subsection{Can the surrogate identify discovery moments?}
\label{sec:discovery-moments}
\begin{figure}
    \centering
    \includegraphics[width=1.0\linewidth]{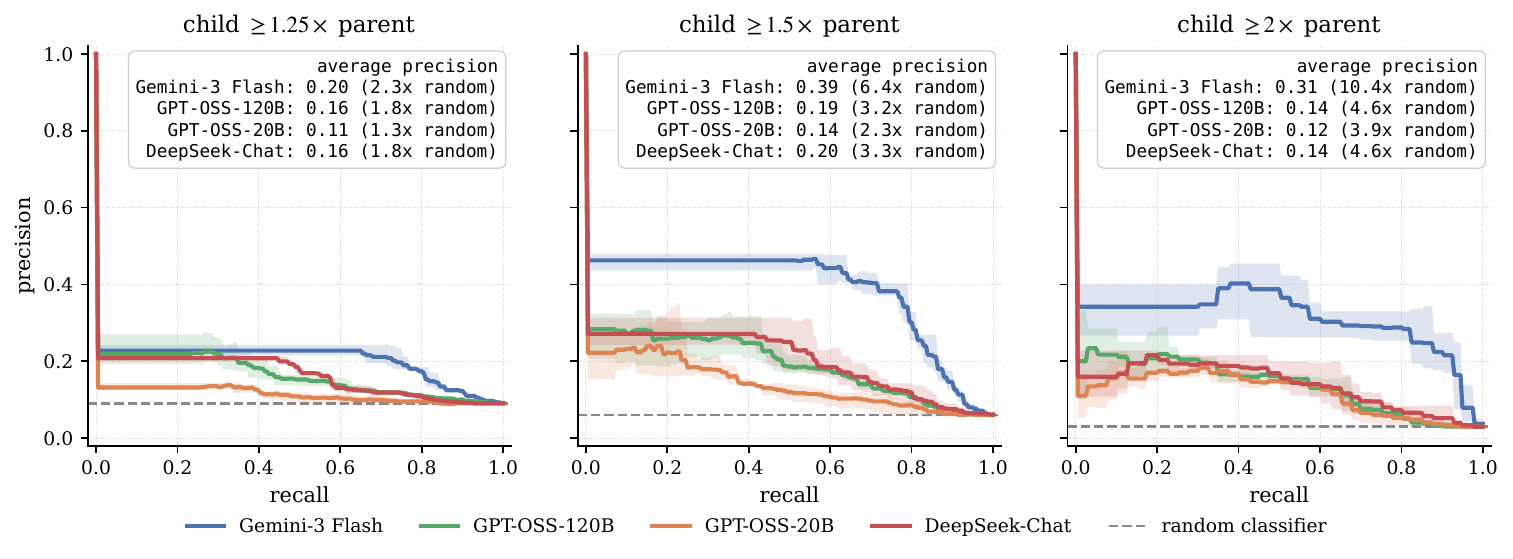}
    \caption{\textbf{Surrogates can identify discovery moments, the parent-to-child mutations where the child is much faster than its parent.}
    Each panel fixes one definition of \emph{much faster} (child at least $1.25\times$, $1.5\times$, or $2\times$ faster than its parent).
    The surrogate ranks all pairs from most to least likely to be a discovery. Each curve shows precision and recall as we accept more of the ranking, starting from just the top most confident pairs on the left and ending with all of them on the right.
    Solid lines are the mean over three repeats, with shaded min/max bands.
    The dashed line is the random-classifier base rate at that threshold.
    }
    \label{fig:discovery-pr}
\end{figure}
How good is the surrogate at telling when a child kernel is much faster than the parent it was mutated from?
We call such a step a \emph{discovery moment}.
These are the steps that break stagnation in a search, so a surrogate that can identify them could prioritize GPU evaluations for the changes that matter most.

The surrogate takes the code of two kernels, $K$ and $K'$, and predicts a distribution over their speedup $g = t(K) / t(K')$, where $t(\cdot)$ is the measured runtime and $g > 1$ means $K'$ is faster.
In the evaluation above, $K$ was the task reference and $K'$ a candidate kernel.
Here $K$ is the parent and $K'$ is the child of a mutation step.

We collect 1{,}347 such pairs from 27 completed PUCT searches, covering 21 problems across the six GPU Mode tasks from \cref{sec:setup} and KernelBench Level 3 problems on A100 and L40S GPUs, keeping every pair in which both kernels ran.
A cutoff $x$ slices these pairs into two classes: a pair is a discovery moment when $g \geq x$, and otherwise it is labeled as a negative. 
This latter class includes steps that change nothing and steps that make the kernel slower.
We use three cutoffs, $x \in \{1.25, 1.5, 2\}$, since no single one is obviously right.

At each cutoff we evaluate the surrogate via a binary classification task.
The surrogate's predicted distribution gives each pair a score $s = \Pr[g \geq x] \in [0, 1]$.
The ground-truth label $y$ is $1$ for the discovery moments and $0$ for the negatives.
A threshold $\tau$ turns the score into a prediction $\hat{y}$, which is $1$ when $s \geq \tau$.
For example, at the cutoff $x = 2$, a child that runs $2.3\times$ faster than its parent has $g = 2.3$ and label $y = 1$.
If the surrogate places $s = 0.7$ of its probability on $g \geq 2$, then any threshold $\tau \leq 0.7$ predicts it correctly.
Lowering $\tau$ predicts a positive label for more pairs, which raises recall and usually lowers precision.
\Cref{fig:discovery-pr} reports the precision-recall curve traced by sweeping $\tau$, one curve per cutoff.

\Cref{fig:discovery-pr} shows the tradeoff between precision and recall for each speedup category, as we vary the threshold for binary classification.
We see that surrogate scores carry useful signal for detecting discovery moments:
at the $1.25\times$ cutoff, the best off-the-shelf surrogate reaches mean average precision $0.20$, where $9\%$ of pairs are positive.
Positives become rarer at stricter cutoffs, but the signal persists.
At the $1.5\times$ cutoff the best mean average precision is $0.39$, where $6\%$ of pairs are positive, and at the $2\times$ cutoff it is $0.31$, where $3\%$ of pairs are positive.
This task is harder than ranking candidates against a fixed task reference because the surrogate must judge an incremental improvement over the search parent rather than absolute quality.
The result is encouraging but not conclusive.
Curves stay above the base rate at every cutoff, 
but precision is not high enough to use the surrogate as a standalone discovery detector without GPU confirmation.

\begin{takeaway}[Discovery moments are detectable but noisy]
Surrogates can identify discovery moments better than chance, especially for larger parent-to-child improvements.
They still produce many false positives: recall can be high, but precision falls quickly, so GPU measurement remains necessary to confirm them.
\end{takeaway}

\section{Related Work}
\paragraph{Kernel optimization and code-execution reasoning.}
Several LLM-driven kernel optimization methods rely on per-candidate device measurement as the search signal, including reinforcement-learned agents~\citep{cuda_agent}, test-time training~\citep{ttt-discover}, and evolutionary loops over strategies~\citep{k_search}, which adapt the broader paradigm of LLM-driven evolutionary program search~\citep{alphaevolve,shinkaevolve,alphaevolve_math,thetaevolve} to kernels.
Each measurement involves compilation, correctness checking, and on-device profiling, which together significantly increase the cost of search.
Other work reduces this cost through domain-specific languages~\citep{cutlass_sol,cutegen}, but every generated candidate is still measured on the device.
Neither direction reduces the per-measurement cost.
We pursue an alternative: studying whether the GPU can be replaced with a surrogate.
This relates to work on the simulatability of code by LLMs, which has examined whether LLMs can predict program input-output behavior~\citep{cruxeval}, simulate step-by-step Python execution~\citep{cwm_meta}, 
simulate unit test outputs \cite{prasadlearning},
and count CUDA kernel FLOPs~\citep{gpuflop_counting} from source.
These results suggest that the same capability could allow LLMs to act as surrogates of physical devices.

\paragraph{World models and surrogate-assisted optimization.}
Several adjacent fields use learned predictors to stand in for an expensive system, reducing the real interactions a downstream procedure requires.
Model-based reinforcement learning learns a predictive model of an environment from interaction~\citep{sutton_dyna,world_models,muzero,dreamerv3}.
Related work asks whether LLMs can learn such models from limited interaction in grid and game environments~\citep{autumnbench,poe_world,onelife}.
Surrogate-assisted optimization uses a learned predictor as a stand-in for an expensive objective during search~\citep{jones_ego}.
Our question is whether we can build a model of a physical GPU, analogous to how prior work has built models of game and puzzle environments or stand-ins for expensive optimization objectives.
We study a learned predictor of a device's runtime response to a kernel, and whether the predictor can identify when its prediction will be accurate.

\paragraph{LLM forecasting, selective prediction, and calibration training.}
LLMs have been evaluated as forecasters of international events~\citep{mirai}, prediction-market resolutions~\citep{prophetarena}, and open-ended forecasting questions about current events~\citep{openforecaster}.
These results show that LLMs carry world knowledge applicable to complex predictions.
Recent work augments RL correctness rewards with a proper scoring rule to train LLMs for calibration~\citep{rlcr}, since LLMs are not by default well-calibrated.
We want such calibration in our setting because it identifies which forecasts can be accepted without measuring every candidate on the GPU.
This connects to selective prediction, where a model is scored on its kept subset under a risk-coverage trade-off~\citep{elyaniv_wiener,geifman_elyaniv}.
In kernel search, sending uncertain candidates to the GPU also resembles active learning, where labels are acquired selectively under a measurement budget.
Our prediction target additionally resembles environment modeling in planning and model-based reinforcement learning.

\section{Conclusion}
GPU measurements are the ground-truth signal for kernel optimization, but they are also a major cost of search.
We studied whether LLMs can reduce this cost by acting as selective surrogates of a physical GPU: forecasting relative kernel speedups, reporting confidence, and reserving device measurement for uncertain cases.
Across held-out Triton kernels, the evaluated surrogates recover most of the available speedup under limited measurement budgets, and their confidence can carry useful information about forecast quality.
Reinforcement learning further shows that the reward matters: calibration-shaped training improves confidence reliability and can improve the ranking signal used by budgeted search, while correctness alone is not enough.
These intrinsic gains carry over to end-to-end search, where a search that uses the surrogate to expand its candidate pool finds faster kernels than an equal-GPU-budget baseline that measures every proposal.
These results suggest a practical path toward kernel optimizers that treat hardware measurement as a scarce resource, using learned GPU surrogates where they are reliable and the device itself where they are not.

\section{Acknowledgments}
This work was supported by NSF-CAREER Award 1846185, NSF AI Engage Institute DRL2112635, Capital One Research Award, NDSEG PhD Fellowship, Microsoft Agentic AI Research and Innovation (AARI) grant program, and a compute grant from Modal. The views contained in this article are those of the authors and not of the funding agency.

\bibliography{references}
\bibliographystyle{plainnat}
\appendix

\section{Success Cases and Failure Cases}
\label{app:qual-cot-grounded-vs-speculative}

We walk through two reasoning traces from the untrained surrogate on TriMul prompts.
The two candidates make similar structural changes to the reference (a candidate-side rewrite of the core triangle product, with bfloat16 or float16 cast).
The measured speedups are very different: the first candidate runs $1.71\times$ faster than the reference, the second runs at parity with the reference ($1.02\times$).
The grounded trace correctly identifies the first as a minor speedup.
The poorly-grounded trace mistakes the second for a more-than-$4\times$ speedup.

\paragraph{Grounded trace, candidate at $1.71\times$ measured speedup, predicted bin $5$.}
The reference applies separate float32 \texttt{nn.Linear} projections, gating sigmoids, and a bfloat16 \texttt{einsum} for the triangle product.
The candidate folds the projections, gating, and product into a float16 path that reuses tensor cores end to end.
The model's full reasoning trace is reproduced here verbatim:

\begin{quote}
\itshape
We need to analyze speedup of candidate relative to reference.
The reference uses torch layers: nn.Linear, LayerNorm etc.
The torch layers under the hood likely use cuBLAS GEMMs for linear operations; the reference uses float32 operations.
The candidate uses FP16 for most computations, with some cast overhead.
It fuses ops etc.
[\ldots] Reference utilizes torch einsum for core TriMul: \texttt{einsum('... i k d, ... j k d -> ... i j d', left.to(torch.bfloat16), right.to(torch.bfloat16))}.
Two input projected to shape $[B, N, N, H]$ then einsum over $k$ dimension ($H$).
That is a batched matrix multiplication: the same dimension as left and right projecting; similar to the candidate's torch.matmul of permuted, I'm not sure if einsum performance similar or better.
Reference uses bfloat16 conversions.
Candidate uses half.
On A100, tensor cores support fp16 and bf16, with similar speeds?
For large matrices maybe BF16 might be slightly faster (because of memory bandwidth?
Actually A100 can handle BF16 and FP16 in Tensor Cores identical).
Candidate uses half for core multiplication too; could yield similar speed.
[\ldots] Overall, candidate likely faster: maybe $1.2\!\times\!\!-\!\!1.4\times$?
On A100 fp16 can be up to $2\times$ faster than fp32 for GEMMs.
But since reference uses bf16 (which also uses Tensor Cores for FP32?
Actually A100 supports BF16 in Tensor Cores; it's similar hardware).
But reference also uses dtype float32 for LayerNorm etc.
[\ldots] Thus likely candidate has FI moderate speedup: $\sim 1.3$.
This puts in bin $5$ (minor speedup).
\end{quote}

The trace is grounded along three lines that each match the actual hardware behavior on A100.
The model recognizes that PyTorch's \texttt{einsum} on a four-dimensional contraction dispatches to a tensor-core matmul, not to a generic semiring kernel.
It correctly identifies that bfloat16 and float16 have similar tensor-core throughput on A100.
It realizes the candidate's win comes from running the surrounding LayerNorm and projections in float16 rather than float32, not from the core triangle product itself.
The numerical estimate that follows ($\sim 1.3\times$) is conservative against the measured $1.71\times$, which reflects model uncertainty rather than incorrect reasoning, and the predicted bin matches the true bin.

\paragraph{Poorly-grounded trace, candidate at $1.02\times$ measured speedup, predicted bin $8$.}
A different candidate on a similar TriMul prompt rewrites the bfloat16 \texttt{einsum} as a permuted batched \texttt{torch.matmul}, leaving the surrounding projections in float32 as in the reference.
The measured candidate runtime is essentially equal to the reference's, but the model places probability $0.65$ on bins $6$, $7$, and $8$ (more than $1.41\times$ speedup), with the argmax on bin $8$ at probability $0.25$ and an additional $0.15$ on bin $5$.
The full reasoning trace:

\begin{quote}
\itshape
We have to compare reference kernel, which uses PyTorch operations including many module calls (Linear, LayerNorm, einstein summation).
The candidate kernel rewires inner einsum into batched matmul which is efficient.
Reference uses einsum which is less efficient, older baseline is naive and may be not optimized by cuBLAS, but PyTorch's einsum uses semiring operations maybe less efficient.
So candidate likely much faster.
How many bits/flops?
For large seq length $N \geq$? Provided config unspecified but typical large can be up to $1024$.
Implementation rewriting ensures using cuBLAS batched gemms.
So expect high speedup maybe $10\times$?
But we have to pick bins up to extreme $>4\times$.
So candidate in bin $8$.
[\ldots] So reference einsum is less efficient than cuBLAS batched matmul.
Also candidate uses torch.matmul which uses batched gemm; it's optimized.
So speedup maybe moderate to high?
Suppose reference $4\times$ slower?
Might be $>4\times$, but maybe $8\times$?
Bin $8$ (extreme) plausible.
[\ldots] Confidence high that candidate faster.
Without measuring, but we expect $>4\times$ due to einsum inefficiency.
\end{quote}

The reasoning collapses on a prior that PyTorch's \texttt{einsum} ``uses semiring operations maybe less efficient'' than \texttt{cuBLAS batched gemm}.
This prior is not supported by the source code in front of the model and is also not supported by the runtime.
The contraction pattern \texttt{... i k d, ... j k d -> ... i j d} is exactly a batched matrix multiplication, and modern PyTorch's \texttt{einsum} dispatches it to the same cuBLAS path that the candidate's \texttt{torch.matmul} does.
Once the model commits to the einsum-is-slow prior it stops looking for confirming evidence.
The trace fabricates a sequence-length default of $N{=}1024$ to motivate a FLOP estimate, declares the candidate ``maybe $8\times$'' faster, and exits to a bin-$8$ argmax at $0.25$ probability with the rest of the mass spread upward toward bin $8$ rather than downward toward bin $4$.
The candidate is in fact at parity with the reference, so the prediction is wrong by four bins, and the source of the error is fully visible in the CoT but not in the short rationale string the surrogate emits to the tool call.

The two traces share the same vocabulary, the same overall structure, and even the same uncertainty about how PyTorch dispatches \texttt{einsum}.
The grounded trace resolves the uncertainty by referencing a checkable property of the hardware (BF16 and FP16 have the same A100 tensor-core throughput) and arrives at a calibrated speedup estimate.
The poorly-grounded trace resolves the same uncertainty by appealing to an unverified prior (``einsum maybe semiring, less efficient'') and arrives at a four-bin error.

\section{Eval-Set Construction and Dataset Compute Footprint}
\label{app:eval-set}

\paragraph{Task selection.}
The six GPU Mode tasks (TriMul, cross-entropy, two Gated DeltaNet forward kernels, GDN-W/U recompute, FP8 quantization) were selected to span operations that appear in current production ML workloads and that have published, human-tuned reference implementations.
Two considerations drove the choice.
First, prior LLM-driven kernel search work~\citep{ttt-discover, k_search} has used the same competition as its evaluation surface, so the operations have an established record as challenging targets for automated kernel optimization.
Second, each task is compositionally non-trivial: it consists of multiple memory and compute stages whose fusion patterns and tile choices materially affect runtime.
We deliberately go deep on a small number of tasks, sweeping a wide range of generated speedups per task, rather than wide across many simpler kernels, so that a surrogate's behavior can be characterized across the full range of speedup outcomes a real kernel optimizer will encounter.

\paragraph{Eval-set assembly.}
For each task, we run the max-reward tree search procedure described in \cref{app:puct} using Gemini-3 Flash and Gemini-3 Pro~\citep{gemini3} as the mutation provider, with the search reward shaped to drive coverage across all eight speedup bins rather than only the fastest tier.
We harvest every successful candidate from each search, measure its runtime against the task's competition reference on an A100, and assign it to one of the eight speedup bins from \cref{tab:speedup-bins}.
We then cap the number of kernels per bin per task to enforce uniform coverage so heavily explored bins do not dominate the loaded view.
The resulting eval set contains 480 kernels: six tasks $\times$ eight bins $\times$ up to ten kernels per bin, with bins approximately uniformly populated subject to availability of correct candidates in each bin.

\paragraph{Training-pool construction.}
The RL training pool is sourced from six GPT-OSS-20B max-reward tree search runs (\cref{app:puct}), one per task, run on A100 hardware and disjoint from the searches that produced the eval set.
We harvest every successful candidate from those six traces, then a pair-construction stage emits three kinds of comparisons used as training rows: \emph{seed-anchored} (each candidate against the task reference), \emph{parent-edit} (each candidate against its parent in the search tree, when both are correct), and \emph{within-pack} (sampled correct-against-correct pairs within a task, stratified by log-speedup magnitude).
Failed candidates do not contribute training rows, and rows are deduplicated by anchor and candidate code with priority seed $>$ parent-edit $>$ within-pack.
Further training-procedure details are in \cref{app:rl-implementation}.

\paragraph{Compute footprint of the underlying searches.}
The kernels in this paper are drawn from a larger archive of 12{,}388 unique LLM-generated GPU kernels with measured runtimes that we release alongside the paper.
The full archive consumed approximately 411M LLM tokens for candidate generation and 618 GPU-hours of A100, L40S, and H100 measurement across all problems and backends.
The portion supplying the GPU Mode held-out eval set used roughly 47 A100-hours of measurement and 62M LLM tokens; the six search runs supplying the RL training pool sit within this archive and represent a small fraction of that total.

\section{Per-Evaluation Latency and Throughput on a Single A100}
\label{app:throughput}

\begin{figure}[t]
    \centering
    \includegraphics[width=0.85\linewidth]{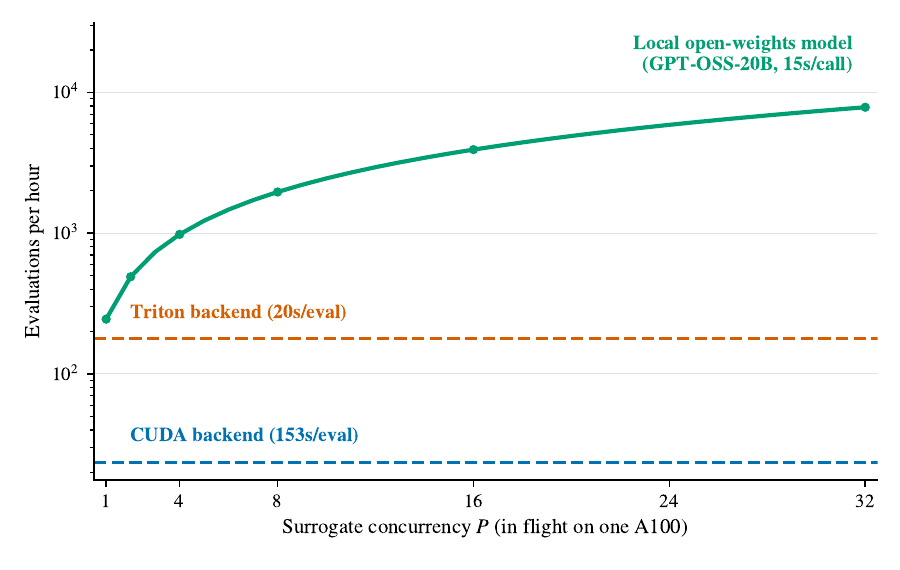}
    \caption{\textbf{Evaluations per hour on a single A100 for the three evaluation modes used in this paper.}
    The two dashed lines are the kernel benchmark, fixed at one in flight because timing a kernel against the reference requires sole occupancy of the GPU.
    The solid line is the LLM surrogate (GPT-OSS-20B), which has no such constraint and serves multiple forecasts concurrently from the same GPU, so its throughput grows with the number $P$ of forecasts in flight.
    Latencies in parentheses are mean per-call wall clock measured during the runs that produced this paper's results.}
    \label{fig:throughput}
\end{figure}

This appendix reports the per-evaluation latency that motivates the surrogate as a low-latency alternative to on-device measurement, and converts that latency into evaluations per hour on a single A100.
\Cref{fig:throughput} summarizes the three evaluation modes used in the paper.

\paragraph{Measurement procedure.}
For each of the three evaluation modes, we recorded the wall clock from the moment a candidate is submitted to the moment its outcome (a measured speedup for the kernel benchmark, a forecast distribution for the surrogate) is returned to the search procedure.
All measurements were taken on a single A100, using the same evaluator implementations the paper uses everywhere else, so the reported latency reflects production conditions rather than a microbenchmark.
We then took the mean across all evaluations performed in the relevant runs, and converted to evaluations per hour by computing $3600 \cdot P / \mathrm{latency}$, where $P$ is the number of evaluations the device serves concurrently.

\paragraph{Three evaluation modes.}
The Triton kernel benchmark compiles each candidate, checks its outputs against the reference, and times it against the reference on the device.
Its mean per-call latency is $20.16$ seconds, averaged over $1{,}515$ evaluations performed across the six GPU Mode kernel-search runs in this paper.
The CUDA kernel benchmark performs the same compile, check, and time steps for hand-written CUDA kernels.
Its mean per-call latency is $152.55$ seconds, averaged over $6{,}446$ evaluations performed across the eleven KernelBench Level 3 search runs in this paper, and is dominated by \texttt{nvcc} compilation rather than on-device execution.
The LLM surrogate runs GPT-OSS-20B locally and returns a calibrated forecast distribution over speedup bins together with a brief chain of reasoning.
Its mean per-call latency is $14.69$ seconds, averaged across the surrogate-scoring run that produced the headline open-weights numbers in the main text.

\paragraph{Concurrency and throughput.}
Throughput depends on how many evaluations a single A100 can serve concurrently.
Timing a kernel against a reference is a measurement of execution time on the device, and is only meaningful when no other workload competes for the GPU, so the kernel benchmark is restricted to one evaluation in flight at a time.
The surrogate has no such requirement: it is a forward pass of an LLM, and the model server hosting GPT-OSS-20B can hold many forecasts in flight simultaneously on the same GPU, so its throughput scales close to linearly with the number of concurrent forecasts $P$ until the device saturates.
Even at $P{=}1$, the surrogate already produces more evaluations per hour than the Triton benchmark, and at modest concurrencies it produces between one and two orders of magnitude more evaluations per hour than the CUDA benchmark.
This is the latency gap that the surrogate exploits: when a forecast is good enough to act on, the search can replace a costly on-device measurement with a much cheaper LLM call, and reserve on-device measurement for the candidates whose forecasts are not trustworthy enough to act on.

\section{Kernel Search via Max-Reward Tree Search}
\label{app:puct}

The runs that produced the held-out eval set and the RL training pool used in this paper share the same outer search procedure: max-reward tree search with a PUCT selection rule~\citep{puct} over an archive of correct candidates.
We refer to it as \emph{max-reward tree search} below, since the backup is over the maximum descendant reward rather than the mean as in textbook MCTS, and PUCT names only the selection score the search uses to choose which archive node to expand next.
This is the same outer loop used by TTT-Discover~\citep{ttt-discover} for kernel optimization.
We retain the search structure and replace TTT-Discover's RL-trained mutation policy with a frontier or open-weights LLM that generates kernel edits from a fixed prompt, so the search behavior is determined by the base model rather than by an additionally trained policy.
The broader kernel archive we release alongside the paper includes runs that used other generation procedures (e.g.\ agentic CLI loops and exploratory variants); those runs are not used in the experiments reported here.

\paragraph{State and statistics.}
The search maintains an archive of correct candidate kernels, each with a measured reward (the candidate's speedup over the task reference).
For every node $s$ in the archive, the search tracks a visit count $n(s)$, the best reward observed across $s$'s children $m(s)$, and a global expansion counter $T$.

\paragraph{Selection score.}
At each step, every archive node is scored by
\[
  \mathrm{score}(s) = Q(s) + c_{\mathrm{puct}}\,\sigma(\mathcal{A})\,P(s)\,\frac{\sqrt{1+T}}{1+n(s)},
\]
where $Q(s) = m(s)$ if $s$ has been expanded and the candidate's own reward otherwise, $\sigma(\mathcal{A})$ is the spread of non-seed rewards in the archive, and $P(s)$ is a rank-based prior over archive rewards normalized to sum to one.
The first term exploits known reward; the second drives exploration toward high-ranked, under-visited nodes and decays as their visit count grows.

\paragraph{Batched expansion with lineage blocking.}
The top-scoring node is expanded each step.
When the search runs with a batch size greater than one, additional parents are added greedily in score order, but no two parents in a single batch may share an ancestor or descendant, so a batch covers distinct branches of the archive.
Each selected parent is sent to the LLM mutation provider, which proposes $K$ candidate edits to the parent's kernel source.
Every proposed candidate is then compiled, correctness-checked, and timed on the target GPU as a single batch evaluation.

\paragraph{Backpropagation and archive maintenance.}
After a batch evaluation, the best valid child reward per parent is folded into $m(\cdot)$ for the direct parent only, and visit counts $n(\cdot)$ are incremented along the parent and all of its ancestors.
$T$ is incremented once per parent expanded.
A parent that produces zero valid children still has its visit counts incremented, so a node that consistently fails has its exploration bonus decay.
Children are inserted into the archive subject to a top-$k$-per-parent cap (by reward), deduplication by program text, and a global capacity bound that preserves seed nodes.

\paragraph{Mutation prompt and reward shaping.}
The LLM mutation provider receives the parent kernel, a description of the target operation, and (in the goal-conditioned configuration used for the eval set) a target speedup bin.
The reward is the geometric-mean speedup over the task's reference inputs, with invalid candidates assigned a sentinel that excludes them from $m$ updates and from archive insertion.
Concrete hyperparameters used in this paper (batch size, samples per parent, $k$ per parent, total step budget, $c_{\mathrm{puct}}$, archive capacity) follow the per-experiment configurations in our released code; the search procedure itself is identical across runs.

\paragraph{Pseudocode.}
\Cref{alg:puct} summarizes the loop.

\begin{algorithm}[t]
\small
\begin{verbatim}
def max_reward_tree_search(seed_program, mutation_provider, evaluation_provider,
                           total_steps, batch_size, samples_per_parent, k_per_parent,
                           c_puct=1.0, capacity=1000):
    root = Node(seed_program, evaluate(seed_program), is_seed=True)
    archive, seed_ids = [root], {root.id}
    n, m, T = {}, {}, 0

    for step in range(total_steps):
        # SELECTION: score every archive node, pick a batch of parents
        # under lineage blocking (no ancestor/descendant pair in one batch).
        scored = puct_scores(archive, n, m, T, seed_ids, c_puct)
        parents = select_with_lineage_blocking(scored, batch_size)
        if not parents:
            break

        # EXPANSION: LLM proposes samples_per_parent edits per parent;
        # all candidates are compiled, correctness-checked, and timed.
        candidates, parent_of = [], []
        for p in parents:
            for code in mutation_provider.mutate(p, samples_per_parent):
                candidates.append(code)
                parent_of.append(p)
        evals = evaluation_provider.batch_evaluate(candidates)
        children = [Node(c, e) for c, e in zip(candidates, evals)]

        # BACKPROPAGATION: m on direct parent, n on parent + ancestors,
        # T += 1 per expanded parent. Failed parents still decay.
        T = backpropagate(children, parent_of, n, m, T)
        for p in parents_with_no_valid_child(parents, children, parent_of):
            T = record_failed_rollout(p, n, T)

        # ARCHIVE UPDATE: top-k children per parent (by reward),
        # dedup by program text, truncate to capacity preserving seeds.
        update_archive(archive, children, parent_of, seed_ids,
                       k_per_parent=k_per_parent, capacity=capacity)

    return archive
\end{verbatim}
\caption{Max-reward tree search with a PUCT selection rule, our primary method to search for kernels. The mutation and evaluation providers are the only components that change between runs: the mutation provider is the LLM that proposes kernel edits, and the evaluation provider compiles, correctness-checks, and times each candidate on the target GPU.}
\label{alg:puct}
\end{algorithm}

\section{Reinforcement Learning Implementation Details}
\label{app:rl-implementation}

All reinforcement-learning variants start from GPT-OSS-20B.
We sample the model with the same reasoning renderer used for the untrained GPT-OSS-20B baseline, so training and scoring share the same prompt format, tool-call schema, and parsing path.

The training set is built from six GPT-OSS-20B kernel-search traces, one for each kernel family used in the paper.
Each row contains an anchor kernel, a candidate kernel, the A100 hardware context, and the candidate's measured speedup relative to the anchor.
Rows are generated from seed-anchored comparisons, parent-edit comparisons, and within-kernel pair comparisons between successful candidates.
Failed candidates are not used for reinforcement learning.

Each rollout is a single-turn interaction.
The prompt contains the reference kernel, the candidate kernel, and hardware properties of the target GPU.
The model samples a chain of reasoning and then calls a structured tool.
In the predict-only variants, this tool returns the predicted speedup bin, a probability distribution over the eight speedup bins, and a short rationale.
We also implemented an exploratory native-abstention variant, in which the model chooses between the forecast tool and a defer tool, but we do not use this variant in the main evaluation.

We train LoRA adapters on GPT-OSS-20B with reinforcement learning.
Each training batch contains groups of rollouts for the same kernel comparison, and the update uses the relative rewards within those groups.
The main text gives the reward definitions; all trained variants use the same training rows and the same training budget unless otherwise noted.
The variants discussed here use twenty reinforcement-learning iterations, group size eight, learning rate $4\times10^{-5}$, rank-$32$ adapters, temperature $1.0$, and a maximum rollout length of $8192$ tokens.

After training, each checkpoint is scored on the canonical held-out evaluation registry using the same forecast schema as the off-the-shelf surrogates.
Each held-out row is sampled three times at temperature $1.0$.
The headline tables compare these scored checkpoints against the same GPT-OSS-20B model before reinforcement learning.

\section{Additional Reproducibility Details}
\label{app:repro-details}

This appendix collects the concrete implementation settings for the main-paper experiments.

\paragraph{GPU measurement protocol.}
We use the same measurement protocol as KernelBench~\citep{kernelbench}.  Each candidate is evaluated independently on a Modal A100-80GB worker.  For each benchmark case, the evaluator sets the Python, NumPy, PyTorch CPU, and PyTorch CUDA RNG seeds to 42, imports the candidate in an isolated temporary module, generates the case input on the GPU, and runs one obligatory correctness pass before timing.  Candidate timing is measured only after the candidate passes correctness.
Timing uses CUDA events with a synchronize before and after every measured call.  The timed function receives a fresh clone of the input data on every iteration, so kernels that mutate their inputs in place do not affect later timing iterations.  The evaluator first times the candidate, then times the reference on the same generated input.  The reported per-case runtime is the mean over an adaptive loop with at least three iterations and at most 100 iterations; the loop stops early when the standard error divided by the mean is below $0.001$, when mean runtime times number of runs exceeds $10$ seconds, or when benchmark wall clock exceeds $120$ seconds.  Per-case speedup is reference mean runtime divided by candidate mean runtime, and the scalar reward used by search and eval-set construction is the geometric mean of per-case speedups across benchmark cases.

\paragraph{Correctness checks and benchmark cases.}
The held-out evaluation set uses six GPU Mode packs, with correctness and benchmark cases taken from the corresponding GPU Mode task definitions.  Correctness cases are separate for TriMul, Gated DeltaNet, and FP8 quantization; cross-entropy uses the same shape family for correctness and runtime with different seeds.  Shape and dtype are checked before numerical closeness for the multi-output packs.  Benchmark cases are the cases used to compute the geometric-mean speedup label.

\begin{itemize}
  \item \textbf{TriMul.}  Correctness uses rtol $2\times10^{-2}$ and atol $2\times10^{-2}$ on 18 public task-specification cases covering normal/Cauchy inputs, mask/no-mask, sequence lengths 32--1024, batch sizes 1--2, dim 128--768, and hidden dim 128.  Benchmark cases are $(N,B,D,H,\mathrm{mask},\mathrm{dist})$: (256,2,128,128,no,normal), (768,1,128,128,no,Cauchy), (256,2,384,128,yes,normal), (512,1,128,128,no,normal), (1024,1,128,128,no,Cauchy), (768,1,384,128,yes,normal), and (1024,1,384,128,no,normal).
  \item \textbf{Cross-entropy.}  Correctness uses rtol $10^{-2}$ and atol $10^{-3}$.  Correctness and benchmark cases both use $B=4096$ with vocabulary sizes 32000, 50264, and 128256.
  \item \textbf{GDN chunk-fwd-h.}  Correctness uses rtol $10^{-3}$ and atol $10^{-3}$ on $(B,T,H,K,V)$ cases (1,64,2,64,64), (2,128,4,64,64), and (1,256,4,64,128).  Benchmark cases are (1,64,1,64,64), (2,512,3,64,64), and (2,1024,3,64,64).
  \item \textbf{GDN chunk-fwd-o.}  Correctness uses rtol $10^{-3}$ and atol $10^{-3}$ with the same Gated DeltaNet correctness and benchmark cases as GDN chunk-fwd-h.
  \item \textbf{GDN recompute-w-u.}  Correctness uses rtol $10^{-3}$ and atol $10^{-3}$ with the same Gated DeltaNet correctness and benchmark cases as GDN chunk-fwd-h.
  \item \textbf{FP8 quantization.}  Correctness uses rtol $10^{-3}$ and atol $10^{-3}$ on $(\mathrm{tokens},D,G)$ cases (1,256,64), (4,512,128), (16,1024,64), (1,4096,128), and (8,4096,128).  Benchmark cases are (256,4096,128), (256,8192,128), and (4096,7168,128).
\end{itemize}

\paragraph{Eval-set construction and search hyperparameters.}
The canonical held-out evaluation set contains the six packs listed above.  For each pack, the evaluation set is capped at 10 kernels per non-failure speedup bin, so the scored evaluation set has at most 80 rows per pack and 480 rows total.  Rows are sorted by bin before capping.  The source searches used to harvest held-out candidates were disjoint from the GPT-OSS-20B searches used to build the RL training rows.

For the source PUCT searches, cross-entropy and GDN chunk-fwd-h were generated with Gemini-3 Flash; GDN chunk-fwd-o, GDN recompute-w-u, and FP8 quantization were generated with Gemini-3 Pro; TriMul was harvested from a separate Gemini-3 Pro A100 search.  The source PUCT searches used one run, 40 budget steps, batch size 2, four samples per parent, top-2 children per parent, archive capacity 1000, $c_{\mathrm{puct}}=1.0$, max LLM concurrency 8, temperature $1.0$, geomean reward aggregation, and A100-80GB evaluation.  The TriMul source search used 10 budget steps, batch size 4, four samples per parent, top-4 children per parent, max LLM concurrency 8, Gemini-3 Pro, A100-80GB evaluation, and geomean aggregation.

Goal-conditioned bin filling used the same setting for every held-out pack: four budget steps, batch size 1, four samples per parent, top-4 children per parent, archive capacity 32, $c_{\mathrm{puct}}=1.0$, 600-second per-request timeout, max LLM concurrency 8, four retries, temperature $1.0$, Triton 3.3.1, A100-SXM4-80GB hardware in the prompt, A100-80GB evaluation, max 10 evaluations in flight, max 10 Modal containers, 1200-second result timeout, and geomean aggregation.  Depending on the pack, bin filling used Gemini-3 Flash, Gemini-3 Pro, or a combination of the two.

\paragraph{RL training and scoring settings.}
The three trained surrogates used in the paper are the final checkpoints from the correctness, correctness+Brier, and correctness+CRPS training runs.  All start from \texttt{openai/gpt-oss-20b} and train LoRA adapters with GRPO through Tinker.  The shared trainer settings are learning rate $4\times10^{-5}$, group size 8, 8 groups per batch, 20 training iterations, checkpoint save interval 5, maximum rollout length 8192 tokens, LoRA rank 32, rollout temperature $1.0$, and the same GPT-OSS-20B prompt/reasoning format used for baseline scoring.  Constant-reward groups are removed before update.

The training set is built from six GPT-OSS-20B source searches over the same packs but disjoint from the held-out evaluation searches.  Rows are successful measurements only and use three anchor sources: seed-anchored rows against the pack reference, parent-edit rows against the successful parent in the PUCT tree, and within-pack successful-candidate pairs.  Within-pack pairs use 500 sampled pairs per pack, RNG seed 0, log-ratio strata $[0,0.1)$, $[0.1,0.5)$, $[0.5,1.0)$, $[1.0,2.0)$, and $[2.0,\infty)$, and no additional tie filtering.

The correctness-only checkpoint uses reward $r=\mathbf{1}[\hat{y}=y]$.  The Brier checkpoint uses $r=\mathbf{1}[\hat{y}=y] + 1-\mathrm{Brier}(p,y)$.  The CRPS checkpoint uses $r=\mathbf{1}[\hat{y}=y] + 1-\mathrm{CRPS}(p,y)$.

\paragraph{Surrogate scoring settings.}
The off-the-shelf Gemini row is scored with LiteLLM model slug \texttt{gemini/gemini-3-flash-preview}; GPT-OSS-120B is scored with \texttt{together\_ai/openai/gpt-oss-120b}; the untrained GPT-OSS-20B row is scored through Tinker with base model \texttt{openai/gpt-oss-20b}.  Gemini and GPT-OSS-120B use temperature $1.0$, max 32000 output tokens, 900-second request timeout, four retries, three repeats per row, and max scoring concurrency 8.  The GPT-OSS-20B baseline and all trained checkpoints use temperature $1.0$, max 16384 output tokens, three repeats per row, max concurrency 8, and the same GPT-OSS-20B prompt/reasoning format.  Each trained scoring run uses the final checkpoint from the corresponding training run and otherwise uses the same scoring pipeline as the untrained GPT-OSS-20B baseline.

\section{Prompts}
\label{app:prompts}

\begin{table}[h!]
  \centering
  \caption{\textbf{Speedup bins used as the surrogate prediction target.}
  The speedup ratio $S_i=T_{\mathrm{ref}}/T_i$ is the candidate runtime improvement relative to the reference implementation, so $S_i>1$ means the candidate is faster and $S_i<1$ means it is slower.
  The representative value $m_b$ is used when converting a bin distribution or predicted bin into a numeric speedup; interior-bin representatives are geometric centers, and open-bin representatives are empirical means over examples in that bin.}
  \label{tab:speedup-bins}
  \small
  \resizebox{\linewidth}{!}{%
  \begin{tabular}{cllll}
    \toprule
    Bin & Name & Speedup range & $m_b$ & Interpretation \\
    \midrule
    1 & Severe slowdown & $S_i \leq 0.25$ & empirical mean & Candidate is at least $4\times$ slower \\
    2 & Significant slowdown & $0.25 < S_i \leq 0.5$ & $0.35$ & Candidate is $2$--$4\times$ slower \\
    3 & Moderate slowdown & $0.5 < S_i \leq 0.71$ & $0.59$ & Candidate is about $1.4$--$2\times$ slower \\
    4 & Minor slowdown & $0.71 < S_i \leq 1.0$ & $0.84$ & Candidate is near baseline but slower \\
    5 & Minor speedup & $1.0 < S_i \leq 1.41$ & $1.19$ & Candidate is slightly faster \\
    6 & Significant speedup & $1.41 < S_i \leq 2.0$ & $1.68$ & Candidate is about $1.4$--$2\times$ faster \\
    7 & High speedup & $2.0 < S_i \leq 4.0$ & $2.83$ & Candidate is $2$--$4\times$ faster \\
    8 & Extreme speedup & $S_i > 4.0$ & empirical mean & Candidate is more than $4\times$ faster \\
    \bottomrule
  \end{tabular}
  }
\end{table}

This appendix contains the exact prompt templates used in the paper.
The surrogate prompts (\cref{box:surrogate-system,box:surrogate-user,box:surrogate-abstain-system,box:surrogate-abstain-user}) are rendered into the LLM forecaster that predicts a candidate kernel's speedup bin.
The mutation prompts (\cref{box:gpu-mode-rules,box:gpu-mode-feedback,box:kernelbench-mutation}) are rendered into the LLM that proposes the next kernel edit during search.
Templates are shown verbatim with their Jinja2 placeholders.
Pack-specific descriptions used by the GPU Mode mutation prompt are not reproduced in full.
We show the rendered structure with a single example (TriMul) to illustrate the format.

\subsection{Surrogate forecaster prompts}

\begin{pabox}[box:surrogate-system]{Surrogate system prompt (predict-only)}
\begin{Verbatim}
You are a GPU execution simulator. Your task is to predict the
relative speedup of a candidate CUDA kernel compared to a reference
kernel implementation. You will reason step-by-step about how each
kernel executes on GPU hardware, then predict which speedup bin the
candidate falls into.

## Speedup Definition

Speedup S = (reference runtime) / (candidate runtime).
- S > 1 means the candidate is FASTER than the reference.
- S < 1 means the candidate is SLOWER than the reference.
- S = 1 means identical performance.

## Speedup Bins

You must predict one of these bins:

| Bin | Name | Speedup Range | Description |
|-----|------|---------------|-------------|
| 1 | SEVERE_SLOWDOWN | S <= 0.25 | Candidate is 4x+ slower |
| 2 | SIGNIFICANT_SLOWDOWN | 0.25 < S <= 0.5 | Candidate is 2x-4x slower |
| 3 | MODERATE_SLOWDOWN | 0.5 < S <= 0.71 | Candidate is ~1.4x-2x slower |
| 4 | MINOR_SLOWDOWN | 0.71 < S <= 1.0 | Candidate is slightly slower or equal |
| 5 | MINOR_SPEEDUP | 1.0 < S <= 1.41 | Candidate is slightly faster |
| 6 | SIGNIFICANT_SPEEDUP | 1.41 < S <= 2.0 | Candidate is ~1.4x-2x faster |
| 7 | HIGH_SPEEDUP | 2.0 < S <= 4.0 | Candidate is 2x-4x faster |
| 8 | EXTREME_SPEEDUP | S > 4.0 | Candidate is 4x+ faster |

## Probability Distribution

Your prediction is a probability distribution over bins 1 through 8.
For each bin, assign a probability in [0, 1] reflecting how likely
the candidate's speedup is to fall in that bin. The eight
probabilities must sum to 1.

Your distribution should reflect genuine uncertainty. For most
predictions, the probability mass is concentrated on 1-2 adjacent
bins (your best guess and a neighbor), with smaller mass on the
rest. Avoid placing all mass on a single bin unless your analysis
is decisive.

## How to Analyze

Simulate the execution of both kernels on the GPU. Consider these
factors systematically:

1. **Algorithmic complexity**: How many FLOPs does each kernel
   perform? Does the candidate reduce total work (e.g., fused
   operations, fewer passes over data)?

2. **Memory access patterns**: Are global memory accesses
   coalesced? Does the candidate use shared memory or registers
   to reduce global memory traffic? How many bytes are read/written
   per thread?

3. **Arithmetic intensity**: What is the ratio of compute to memory
   operations? Is the kernel compute-bound or memory-bandwidth-
   bound? This determines which optimizations matter.

4. **Thread divergence**: Do conditionals cause warp divergence?
   Does the candidate reduce branch divergence compared to the
   reference?

5. **Occupancy and resource pressure**: How many registers per
   thread? How much shared memory per block? These limit the
   number of concurrent warps and can bottleneck throughput.

6. **Parallelism and grid dimensions**: Does the candidate expose
   more parallelism? Are there enough threads to saturate the GPU?
   Is the work evenly distributed across blocks?

7. **Synchronization overhead**: Does the kernel use
   __syncthreads(), atomics, or other synchronization primitives?
   These can serialize execution.

8. **Kernel launch overhead**: For very fast kernels, launch
   overhead can dominate. Multiple kernel launches vs. a single
   fused kernel matters.

9. **Library calls vs. custom code**: Does the reference use highly
   optimized vendor libraries (cuBLAS, cuDNN)? Custom kernels
   rarely beat these for standard operations (GEMM, convolution).

10. **Data types and special hardware**: Does the candidate use
    lower precision (FP16, INT8) or tensor cores? These can
    provide large speedups for eligible operations.

Think carefully about the RELATIVE performance. A candidate might
be well-written but still slower than a PyTorch reference that
dispatches to cuBLAS. Conversely, a simple-looking candidate might
achieve large speedups by fusing operations that the reference
executes as separate kernel launches.

## Output Format

Reason step by step about both kernels, then submit your prediction
by calling the tool `{{ tool_name }}` exactly **once**. Do not
respond in plain text and do not call any other tool.

The tool's arguments are:
- `predicted_bin`: integer in [1, 8], the most likely bin.
- `p_severe_slowdown` ... `p_extreme_speedup`: eight floats in
  [0, 1] that sum to 1, one per bin in order.
- `reasoning`: 1-3 sentence rationale for the prediction.
\end{Verbatim}
\end{pabox}

\begin{pabox}[box:surrogate-user]{Surrogate user prompt (predict-only)}
\begin{Verbatim}
## Task: {{ task.op_name }} (Level {{ task.level_id }})

Predict the speedup of the candidate kernel relative to the
reference kernel.

## Reference Kernel: {{ reference.kernel_name }}

```python
{{ reference.code }}
```

## Candidate Kernel: {{ candidate.kernel_name }}

```cuda
{{ candidate.code }}
```
{% if hardware %}

## Hardware Context

| Property | Value |
|----------|-------|
| Device | {{ hardware.device_name }} |
| Compute Capability | {{ hardware.compute_capability[0] }}.{{ hardware.compute_capability[1] }} |
| Global Memory | {{ "%.1f"|format(hardware.total_global_memory_gb) }} GB |
| Multiprocessors | {{ hardware.multiprocessor_count }} |
| Max Threads/SM | {{ hardware.max_threads_per_multiprocessor }} |
| Clock Rate | {{ "%.2f"|format(hardware.clock_rate_ghz) }} GHz |
| Memory Clock | {{ "%.2f"|format(hardware.memory_clock_rate_ghz) }} GHz |
| Memory Bus Width | {{ hardware.memory_bus_width_bits }} bits |
{% endif %}

Analyze both kernels and predict the speedup bin. Submit your
prediction by calling the `{{ tool_name }}` tool exactly once.
\end{Verbatim}
\end{pabox}

\begin{pabox}[box:surrogate-abstain-system]{Surrogate system prompt (predict-or-defer variant)}
\begin{Verbatim}
You are a GPU execution simulator. Your task is to either predict
the relative speedup of a candidate CUDA kernel compared to a
reference kernel implementation, or, when you are too uncertain to
predict reliably, defer the candidate to a real GPU evaluator. You
will reason step-by-step about how each kernel executes on GPU
hardware, then either submit a probability distribution over
speedup bins or call the deferral tool.

## Speedup Definition

Speedup S = (reference runtime) / (candidate runtime).
- S > 1 means the candidate is FASTER than the reference.
- S < 1 means the candidate is SLOWER than the reference.
- S = 1 means identical performance.

## Speedup Bins

When you predict, you must select one of these bins as the most
likely outcome:

| Bin | Name | Speedup Range | Description |
|-----|------|---------------|-------------|
| 1 | SEVERE_SLOWDOWN | S <= 0.25 | Candidate is 4x+ slower |
| 2 | SIGNIFICANT_SLOWDOWN | 0.25 < S <= 0.5 | Candidate is 2x-4x slower |
| 3 | MODERATE_SLOWDOWN | 0.5 < S <= 0.71 | Candidate is ~1.4x-2x slower |
| 4 | MINOR_SLOWDOWN | 0.71 < S <= 1.0 | Candidate is slightly slower or equal |
| 5 | MINOR_SPEEDUP | 1.0 < S <= 1.41 | Candidate is slightly faster |
| 6 | SIGNIFICANT_SPEEDUP | 1.41 < S <= 2.0 | Candidate is ~1.4x-2x faster |
| 7 | HIGH_SPEEDUP | 2.0 < S <= 4.0 | Candidate is 2x-4x faster |
| 8 | EXTREME_SPEEDUP | S > 4.0 | Candidate is 4x+ faster |

## Probability Distribution

When you predict, you produce a probability distribution over bins
1 through 8. For each bin, assign a probability in [0, 1]
reflecting how likely the candidate's speedup is to fall in that
bin. The eight probabilities must sum to 1.

Your distribution should reflect genuine uncertainty. For most
predictions, the probability mass is concentrated on 1-2 adjacent
bins (your best guess and a neighbor), with smaller mass on the
rest. Avoid placing all mass on a single bin unless your analysis
is decisive.

## How to Analyze

Simulate the execution of both kernels on the GPU. Consider these
factors systematically:

1. **Algorithmic complexity**: How many FLOPs does each kernel
   perform? Does the candidate reduce total work (e.g., fused
   operations, fewer passes over data)?

2. **Memory access patterns**: Are global memory accesses
   coalesced? Does the candidate use shared memory or registers
   to reduce global memory traffic? How many bytes are read/written
   per thread?

3. **Arithmetic intensity**: What is the ratio of compute to memory
   operations? Is the kernel compute-bound or memory-bandwidth-
   bound? This determines which optimizations matter.

4. **Thread divergence**: Do conditionals cause warp divergence?
   Does the candidate reduce branch divergence compared to the
   reference?

5. **Occupancy and resource pressure**: How many registers per
   thread? How much shared memory per block? These limit the
   number of concurrent warps and can bottleneck throughput.

6. **Parallelism and grid dimensions**: Does the candidate expose
   more parallelism? Are there enough threads to saturate the GPU?
   Is the work evenly distributed across blocks?

7. **Synchronization overhead**: Does the kernel use
   __syncthreads(), atomics, or other synchronization primitives?
   These can serialize execution.

8. **Kernel launch overhead**: For very fast kernels, launch
   overhead can dominate. Multiple kernel launches vs. a single
   fused kernel matters.

9. **Library calls vs. custom code**: Does the reference use highly
   optimized vendor libraries (cuBLAS, cuDNN)? Custom kernels
   rarely beat these for standard operations (GEMM, convolution).

10. **Data types and special hardware**: Does the candidate use
    lower precision (FP16, INT8) or tensor cores? These can
    provide large speedups for eligible operations.

Think carefully about the RELATIVE performance. A candidate might
be well-written but still slower than a PyTorch reference that
dispatches to cuBLAS. Conversely, a simple-looking candidate might
achieve large speedups by fusing operations that the reference
executes as separate kernel launches.

## Predict or Defer

You have two tools available and must call exactly one of them:

- **`{{ predict_tool_name }}`** -- submit your speedup-bin
  distribution as a prediction.
- **`{{ defer_tool_name }}`** -- defer this candidate to a real GPU
  evaluator instead of predicting.

Use **`{{ predict_tool_name }}`** when your analysis converges on a
small set of plausible bins. It is acceptable for the distribution
to span two or three adjacent bins; that is what the probability
simplex is for.

Use **`{{ defer_tool_name }}`** when, after analysis, you would not
bet on your top bin being correct. Concretely, defer when:

- your analysis leaves two or more bins close enough that you
  cannot order them confidently, and they span a range wide enough
  that getting the top bin wrong is meaningful (e.g., a 2x or
  larger speedup range);
- the candidate uses an optimization or hardware feature whose
  effect on this specific reference cannot be reasoned about from
  the code alone (e.g., the runtime depends on cache behavior or on
  a tunable parameter whose value matters more than the algorithm);
- the candidate has multiple plausible failure modes (e.g., it
  might be incorrect, or it might compile but produce numerically
  unstable results) and these modes would land it in different
  bins;
- the reference is a vendor-library dispatch and you cannot tell
  whether the candidate is competitive with that library on this
  hardware.

Deferral is not free. In deployment, a deferred candidate is sent
to a real GPU for evaluation, which is expensive and slow. Reserve
deferral for the candidates where you genuinely would not bet on
your prediction. Predicting confidently and being right is the best
outcome; predicting confidently and being wrong is the worst;
deferring is a middle ground when you cannot tell.

## Output Format

Reason step by step about both kernels. Then call exactly one tool:
either **`{{ predict_tool_name }}`** with your bin distribution, or
**`{{ defer_tool_name }}`** with a brief reason. Do not respond in
plain text, and do not call both tools.

Arguments of `{{ predict_tool_name }}`:
- `predicted_bin`: integer in [1, 8], the most likely bin.
- `p_severe_slowdown` ... `p_extreme_speedup`: eight floats in
  [0, 1] that sum to 1, one per bin in order.
- `reasoning`: 1-3 sentence rationale for the prediction.

Arguments of `{{ defer_tool_name }}`:
- `reason`: 1-2 sentences naming the specific source of uncertainty
  that motivates deferral.
\end{Verbatim}
\end{pabox}

\begin{pabox}[box:surrogate-abstain-user]{Surrogate user prompt (predict-or-defer variant)}
\begin{Verbatim}
## Task: {{ task.op_name }} (Level {{ task.level_id }})

Either predict the speedup of the candidate kernel relative to the
reference kernel, or defer this candidate to a real GPU evaluator
if you are too uncertain to predict reliably.

## Reference Kernel: {{ reference.kernel_name }}

```python
{{ reference.code }}
```

## Candidate Kernel: {{ candidate.kernel_name }}

```cuda
{{ candidate.code }}
```
{% if hardware %}

## Hardware Context

| Property | Value |
|----------|-------|
| Device | {{ hardware.device_name }} |
| Compute Capability | {{ hardware.compute_capability[0] }}.{{ hardware.compute_capability[1] }} |
| Global Memory | {{ "%.1f"|format(hardware.total_global_memory_gb) }} GB |
| Multiprocessors | {{ hardware.multiprocessor_count }} |
| Max Threads/SM | {{ hardware.max_threads_per_multiprocessor }} |
| Clock Rate | {{ "%.2f"|format(hardware.clock_rate_ghz) }} GHz |
| Memory Clock | {{ "%.2f"|format(hardware.memory_clock_rate_ghz) }} GHz |
| Memory Bus Width | {{ hardware.memory_bus_width_bits }} bits |
{% endif %}

Analyze both kernels. Then call exactly one of the two tools:
`{{ predict_tool_name }}` to submit a prediction, or
`{{ defer_tool_name }}` if you are too uncertain to predict.
\end{Verbatim}
\end{pabox}

\subsection{Mutation prompts}

The GPU Mode mutation prompt is assembled from a per-kernel description body, a fixed rules block, and a feedback section that depends on the previous candidate's evaluation outcome.
\Cref{box:gpu-mode-rules} shows the rules block appended after every per-kernel description, and \cref{box:gpu-mode-feedback} shows the feedback section appended once a previous candidate has been evaluated.
The kernel description body is pack-specific.
\Cref{box:gpu-mode-trimul-desc} shows the body for the TriMul kernel as a representative example.
\Cref{box:kernelbench-mutation} shows the KernelBench mutation prompt, which uses a single Jinja2 template covering the seed and feedback cases.

\begin{pabox}[box:gpu-mode-rules]{GPU Mode mutation prompt: rules block (appended to every per-kernel description)}
\begin{Verbatim}
Rules:
- The tensors arguments passed in will be already on your cuda device.
- Define all of your code in one final ```python ``` block.
- We will test the correctness of your kernel on multiple input shapes, make sure to support different potential test cases.
- You are allowed to use mixed precision computations, but make sure your final output is in float32.
- You must use triton {triton_version} and these kernels will be run on an Nvidia {gpu_name}.
- You do not have to implement everything in triton, you may choose to have some of the operations done in pytorch. However, you must implement at least part of the operations in a kernel.
- Include a short docstring at the top summarizing your algorithm.
\end{Verbatim}
\end{pabox}

\begin{pabox}[box:gpu-mode-feedback]{GPU Mode mutation prompt: feedback section (appended after a previous candidate has been evaluated)}
\begin{Verbatim}
<base prompt>

Here is your latest implementation:
```python
<previous_kernel_code>
```

Your custom_kernel was evaluated on GPU.

Here is the evaluation result:
<one of the four arms below>

# CompileFailed arm
Your kernel failed to compile.

Compilation error:
<compilation_error, head-truncated to 2000 chars>

Please fix the errors and try again.

# RuntimeError arm
Your kernel raised an exception at runtime.

Error type: <runtime_error_name>

Error message:
<runtime_error, head-truncated to 1000 chars>

Traceback:
<traceback, tail-truncated to 3000 chars>

Please fix the errors and try again.

# Incorrect arm
Your kernel produced incorrect output compared to the reference.

Correctness issue:
<error_message, head-truncated to 2000 chars>

Please fix the correctness issues and try again.

# Success arm
You are iteratively optimizing runtime (microseconds).

Your kernel is correct. Aggregated speedup: <speedup>x (aggregation method: <method>).

Per-case breakdown (slowest first):
  <pack-specific case formatting, one line per test case>
  ...

Please rewrite the entire kernel to be as fast as possible. Focus on the slowest configurations listed above.
\end{Verbatim}
\end{pabox}

\begin{pabox}[box:gpu-mode-trimul-desc]{GPU Mode mutation prompt: kernel description body (TriMul example)}
\begin{Verbatim}
You are an expert Triton engineer tasked with translating PyTorch code into highly optimized Triton kernel code.

You will be implementing a Triangle Multiplicative Update (TriMul) module that is a core operation
for AlphaFold3, Chai, Protenix, and other protein structure prediction models in BioML.

The TriMul operator operates over a 4D tensor of shape [B, N, N, C].

Your task:
- Implement the "outgoing" version of the TriMul operator from the AlphaFold3 paper.
- You will not have to compute or store gradients for this version. You will only need to implement the forward pass.

Your function should be defined as 'custom_kernel' with the following signature:
Input:
- `data`: Tuple of (input: torch.Tensor, weights: Dict[str, torch.Tensor], config: Dict)
    - input: Input tensor of shape [bs, seq_len, seq_len, dim]
    - mask: Mask tensor of shape [bs, seq_len, seq_len]
    - weights: Dictionary containing model weights
    - config: Dictionary containing model configuration parameters

Output:
- output: Processed tensor [bs, seq_len, seq_len, dim]

**Problem Constraints:**
- B in {1,2}, N in {128,256,512,1024}, c in {128}, c_z in {128,384,768}
- The input distribution will be sampled from a standard Normal distribution, or a heavy-tailed Cauchy distribution (gamma = 2).
- There will either be no mask, or a randomly sampled mask over the inputs.

**Remarks.** So why is this problem so annoying? Because you have to choose whether to load / deal with either the channel dimensions c,c_z that the LayerNorms require (otherwise you have to do a synchronize to compute the statistics like mean / variance) or the sequence dimension N.
The sequence dimension is particularly annoying because it's quite large, but also because we compute pair-wise operations at the last operation that sum over another sequence dimension (this is N^3!).
However, I really like this kernel because it only consists of "simple" operations, and is really easy to understand. It is a true test of "fusions" that torch.compile() doesn't do that well.

Here is a pytorch implementation of the TriMul module. You will want to implement a kernel for the operations in the forward call:

```python
import torch
from torch import nn, einsum
import math

# Reference code in PyTorch
class TriMul(nn.Module):
    def __init__(self, dim: int, hidden_dim: int):
        super().__init__()
        self.norm = nn.LayerNorm(dim)
        self.left_proj = nn.Linear(dim, hidden_dim, bias=False)
        self.right_proj = nn.Linear(dim, hidden_dim, bias=False)
        self.left_gate = nn.Linear(dim, hidden_dim, bias=False)
        self.right_gate = nn.Linear(dim, hidden_dim, bias=False)
        self.out_gate = nn.Linear(dim, hidden_dim, bias=False)
        self.to_out_norm = nn.LayerNorm(hidden_dim)
        self.to_out = nn.Linear(hidden_dim, dim, bias=False)

    def forward(self, x: torch.Tensor, mask: torch.Tensor) -> torch.Tensor:
        batch_size, seq_len, _, dim = x.shape
        x = self.norm(x)
        left = self.left_proj(x)
        right = self.right_proj(x)
        mask = mask.unsqueeze(-1)
        left = left * mask
        right = right * mask
        left_gate = self.left_gate(x).sigmoid()
        right_gate = self.right_gate(x).sigmoid()
        out_gate = self.out_gate(x).sigmoid()
        left = left * left_gate
        right = right * right_gate
        out = einsum('... i k d, ... j k d -> ... i j d', left, right)
        out = self.to_out_norm(out)
        out = out * out_gate
        return self.to_out(out)
```

Here is some example skeleton code of the entrypoint function you will create:
```python
def custom_kernel(data):
    input_tensor, mask, weights, config = data
    dim, hidden_dim = config["dim"], config["hidden_dim"]
    norm_weight = weights["norm.weight"]
    norm_bias = weights["norm.bias"]
    left_proj_weight = weights["left_proj.weight"]
    right_proj_weight = weights["right_proj.weight"]
    left_gate_weight = weights["left_gate.weight"]
    right_gate_weight = weights["right_gate.weight"]
    out_gate_weight = weights["out_gate.weight"]
    to_out_norm_weight = weights["to_out_norm.weight"]
    to_out_norm_bias = weights["to_out_norm.bias"]
    to_out_weight = weights["to_out.weight"]
    # Perform TriMul
    return out
```

A few general triton tips:
- tl.arange only takes in constexpr arguments (static or tl.constexpr)
- You cannot use continue in your kernel code
- tl.dot can only take in two input tensors
- There is no tl.mean

Here are the different configs that your kernel will be tested on ("nomask" sets whether there will be no mask, or a randomly sampled mask over the inputs):

Test Cases for correctness and runtime (optimize runtime for these):
  - {"seqlen": 256, "bs": 2, "dim": 128, "hidden_dim": 128, "nomask": True, "distribution": "normal"}
  - {"seqlen": 768, "bs": 1, "dim": 128, "hidden_dim": 128, "nomask": True, "distribution": "cauchy"}
  - {"seqlen": 256, "bs": 2, "dim": 384, "hidden_dim": 128, "nomask": False, "distribution": "normal"}
  - {"seqlen": 512, "bs": 1, "dim": 128, "hidden_dim": 128, "nomask": True, "distribution": "normal"}
  - {"seqlen": 1024, "bs": 1, "dim": 128, "hidden_dim": 128, "nomask": True, "distribution": "cauchy"}
  - {"seqlen": 768, "bs": 1, "dim": 384, "hidden_dim": 128, "nomask": False, "distribution": "normal"}
  - {"seqlen": 1024, "bs": 1, "dim": 384, "hidden_dim": 128, "nomask": True, "distribution": "normal"}
\end{Verbatim}
\end{pabox}

\begin{pabox}[box:kernelbench-mutation]{KernelBench mutation prompt}
\begin{Verbatim}
You are optimizing a PyTorch module by writing custom CUDA kernels.

# Task

Below is a PyTorch module called `Model`. Your job is to produce a new module called `ModelNew` that computes the same outputs but runs faster on GPU. The mechanism is **writing custom CUDA kernels** and compiling them inline with `torch.utils.cpp_extension.load_inline`, then using them inside `ModelNew.forward`. You decide which operations to target: fuse and replace where it pays back, and leave the rest as ordinary PyTorch (e.g. keep a large `torch.mm` since cuBLAS is hard to beat). **`ModelNew` must call at least one custom CUDA kernel that you write yourself** -- pure-PyTorch rewrites do not count, even if they happen to be faster.

Reference module:

```python
{{ reference_module }}
```
{% if feedback is not none %}

# Previous attempt

Here is your previous `ModelNew`:

```python
{{ previous_kernel_code }}
```

It was evaluated on GPU. Result:
{% if feedback.kind == "compile_failed" %}

**Compilation failed.**
{% if feedback.compilation_error_name %}

Error type: `{{ feedback.compilation_error_name }}`
{% endif %}

Compiler output:

```
{{ compilation_error_truncated }}
```

Fix the compilation error. Keep the rest of the design unless the fix requires changing it.
{% elif feedback.kind == "runtime_error" %}

**Runtime error during execution.**
{% if feedback.runtime_error_name %}

Error type: `{{ feedback.runtime_error_name }}`
{% endif %}

Message:

```
{{ runtime_error_truncated }}
```

Traceback:

```
{{ runtime_error_traceback_truncated }}
```

Fix the runtime error. Common culprits: out-of-bounds indexing, dtype mismatch, missing `.contiguous()`, a kernel launch grid that does not cover all elements, or a CUDA-side assertion such as a misaligned address.
{% elif feedback.kind == "incorrect" %}

**Output does not match the reference.**

Issue: {{ correctness_issue_truncated }}
{% if feedback.max_difference %}

Max absolute differences observed across test cases: {{ feedback.max_difference }}
{% endif %}
{% if feedback.avg_difference %}
Average absolute differences observed across test cases: {{ feedback.avg_difference }}
{% endif %}

Make the output match the reference within standard tolerance (atol=1e-2, rtol=1e-2). Do not change the architecture beyond what is required to fix correctness.
{% elif feedback.kind == "success" %}

**Correct.** Speedup: {{ "%.3f"|format(feedback.speedup) }}x (kernel = {{ "%.1f"|format(feedback.runtime_us) }} us, reference = {{ "%.1f"|format(feedback.ref_runtime_us) }} us).

Now make it faster. Rewrite the entire `ModelNew` from scratch with a more aggressive optimization strategy. Do not tune the existing kernel incrementally -- propose a different approach (e.g. fuse more operations across the forward pass, change the tiling, switch to a more coalesced memory access pattern, or use cuBLAS where it beats handwritten CUDA).
{% endif %}
{% endif %}

# Engineering guidance

Treat correctness as a hard prerequisite and speed as the objective.

- **Pick replacements that pay back.** Fusing several small element-wise ops, replacing a launch-bound sequence, or making memory accesses more coalesced tends to win. Reimplementing a single large GEMM in handwritten CUDA rarely beats cuBLAS.
- **Precision.** Inputs are `float32` on a CUDA device. Your final output must be `float32`. Mixed-precision inside is fine if the result still matches the reference.
- **Shapes.** Test shapes are determined by `get_inputs()` and `get_init_inputs()` in the reference module above. Size your kernels around what those imply. Do not hardcode shapes that those functions do not fix.
- **Compile flags.** You may pass `extra_cuda_cflags=["-O3"]` (or similar) to `load_inline`. Do not assume non-default headers.
- **Isolation.** Your code is `exec`'d in an isolated namespace. No filesystem access, no network, no `__file__`.

# How to ship code

Your final code is loaded via `torch.utils.cpp_extension.load_inline` and instantiated as `ModelNew`. For this to work:

- Define a class named exactly **`ModelNew`** that subclasses `torch.nn.Module`. It is the entry point.
- Compile your CUDA source(s) with `torch.utils.cpp_extension.load_inline(...)` at module scope (top-level), not inside `__init__`.
- `ModelNew.__init__` must accept the same arguments as `Model.__init__` (whatever `get_init_inputs()` returns).
- `ModelNew.forward` must accept the same arguments and return the same shape and dtype as `Model.forward`.

A minimal example of the skeleton -- pattern only, not problem-relevant:

```python
{{ skeleton_example }}
```

# Output format

You may think through the design above as much as you like. Put your **final, complete `ModelNew` module in a single ```python``` code block at the end of your response**. The grader extracts the last python code block; everything above it is ignored.
\end{Verbatim}
\end{pabox}

\section{Licenses for External Assets}
\label{app:licenses}

This appendix lists the third-party assets used in the paper, their sources, and their licenses.
Each asset is also cited at first use in the main text.

\paragraph{GPU Mode kernel optimization competition.}
Reference kernels and task specifications are drawn from the GPU Mode leaderboard at \url{https://www.gpumode.com/}.
The supporting kernel-bot infrastructure (\url{https://github.com/gpu-mode/kernelbot}) is released under the MIT License.
Individual reference implementations are contributed by the community and used here under their respective open-source licenses as indicated on the leaderboard.

\paragraph{KernelBench.}
KernelBench~\citep{kernelbench} is released by the Scaling Intelligence Lab at Stanford under the MIT License (\url{https://github.com/ScalingIntelligence/KernelBench}).
We use the published Level 3 task set without modification to the reference implementations.

\paragraph{Sakana AI CUDA Engineer Archive.}
The AI CUDA Engineer Archive is released by Sakana AI on Hugging Face (\url{https://huggingface.co/datasets/SakanaAI/AI-CUDA-Engineer-Archive}) under the CC-BY-4.0 license.
We refer to it as an external comparison point for kernel-archive scale.

\paragraph{Models.}
GPT-OSS-20B is released under the Apache-2.0 license.
Gemini 3 Flash and Gemini 3 Pro are accessed through the Google AI API under its standard terms of service.
DeepSeek models are used under the DeepSeek License.

\section{Example Eval-Set Kernels}
\label{app:example-kernels}

This appendix shows two kernels drawn from the FP8 per-token-group quantization eval set, one near the lower end of the speedup distribution and one near the upper end.
Both kernels target the same operation against the same A100 reference implementation, so the comparison isolates implementation quality rather than task or hardware difficulty.
We selected these examples from the held-out eval set described in \cref{app:eval-set} as representative cases that a surrogate is asked to forecast.

\paragraph{Low-speedup candidate ($S_i = 0.996$).}
\Cref{box:fp8-slow} reshapes the input into per-token groups, computes the per-group absolute maximum, derives the scale, and writes the quantized values back through a sequence of unfused PyTorch tensor operations.
Each step launches its own kernel and materializes an intermediate tensor, so per-element memory traffic dominates and the candidate runs slightly slower than the reference.
The implementation is correct and otherwise unremarkable, which makes it a representative example of the naive end of what an LLM proposes before any fusion or layout work.

\paragraph{High-speedup candidate ($S_i = 1.970$).}
\Cref{box:fp8-fast} fuses the absmax reduction, scale derivation, and clamped quantization into a single Triton launch.
The host-side wrapper picks \texttt{BLOCK\_GROUPS} so that the dispatch grid is deep enough to saturate the streaming multiprocessors of an A100 while bounding per-thread register pressure, and it precomputes the reciprocal of the per-group scale so the elementwise quantization issues a multiplication rather than a division.
The reduction over each group is performed inside registers without a shared-memory round trip.
Compared to the low-speedup candidate, this implementation removes both the unfused intermediate tensors and the per-step launch overhead, which accounts for the measured $1.97\times$ speedup over the reference.

\begin{pabox}[box:fp8-slow]{Low-speedup FP8 quantization candidate}
Measured speedup $S_i=0.996$ on A100.
\begin{Verbatim}
import torch


FP8_MAX = 448.0
FP8_MIN = -448.0
FP8_EPS = 1e-10


def custom_kernel(data):
    """Pure-PyTorch per-token-group FP8 quantization.

    Args:
        data: tuple ``(x, x_q, x_s)`` where
            - x:   [num_tokens, hidden_dim]                float32 on CUDA --- input
            - x_q: [num_tokens, hidden_dim]                float32 on CUDA --- pre-allocated output buffer
            - x_s: [num_tokens, hidden_dim // group_size]  float32 on CUDA --- pre-allocated scale buffer

    Returns:
        Tuple ``(x_q, x_s)`` with the same buffers populated in-place.
    """
    x, x_q, x_s = data
    num_tokens, hidden_dim = x.shape
    num_groups = x_s.shape[1]
    group_size = hidden_dim // num_groups

    x_f32 = x.float()
    x_grouped = x_f32.reshape(num_tokens, num_groups, group_size)

    absmax = x_grouped.abs().amax(dim=-1).clamp(min=FP8_EPS)
    scale = absmax / FP8_MAX

    quantized = (x_grouped / scale.unsqueeze(-1)).clamp(FP8_MIN, FP8_MAX)
    quantized = quantized.reshape(num_tokens, hidden_dim)

    x_q[...] = quantized
    x_s[...] = scale
    return x_q, x_s
\end{Verbatim}
\end{pabox}

\begin{pabox}[box:fp8-fast]{High-speedup FP8 quantization candidate}
Measured speedup $S_i=1.970$ on A100.
\begin{Verbatim}
"""
Optimized Fused Per-Token-Group FP8 (E4M3) Quantization Kernel.

Algorithm:
1. Translates the 2D tensor `(num_tokens, hidden_dim)` into a flat 1D grid of `total_groups`.
2. Dynamically adjusts `BLOCK_GROUPS` strictly to balance SM wave occupancy and register limit.
   - Targeting >= 512 blocks ensures massive wave overlap on A100's 108 SMs to natively hide DRAM latency.
   - Capping `BLOCK_GROUPS` at 128 guarantees <= 64 elements per thread, completely eliminating
     register spilling which otherwise drastically hurts performance on large shapes.
3. Groups are processed entirely within registers natively (intra-warp reduction), bypassing
   shared memory round-trips.
4. Drops PTX masking globally when the dispatch grid perfectly overlaps the tensor boundary.
"""

import torch
import triton
import triton.language as tl

@triton.jit
def fp8_quantize_kernel(
    x_ptr, x_q_ptr, x_s_ptr,
    total_groups,
    GROUP_SIZE: tl.constexpr,
    BLOCK_GROUPS: tl.constexpr,
    NEEDS_MASK: tl.constexpr,
):
    pid = tl.program_id(0)
    group_start = pid * BLOCK_GROUPS

    group_offsets = group_start + tl.arange(0, BLOCK_GROUPS)
    elem_offsets = tl.arange(0, GROUP_SIZE)

    # [BLOCK_GROUPS, GROUP_SIZE] layout ensures vectorized load instructions (ld.global.v4)
    offsets = group_offsets[:, None] * GROUP_SIZE + elem_offsets[None, :]

    if NEEDS_MASK:
        mask_1d = group_offsets < total_groups
        mask = mask_1d[:, None]
        x = tl.load(x_ptr + offsets, mask=mask, other=0.0)
    else:
        x = tl.load(x_ptr + offsets)

    # Intra-warp reduction over elements (no shared memory needed as GROUP_SIZE fits in local thread registers)
    absmax = tl.max(tl.abs(x), axis=1)
    absmax = tl.maximum(absmax, 1e-10)

    scale = absmax / 448.0
    inv_scale = 448.0 / absmax

    x_q = x * inv_scale[:, None]
    x_q = tl.minimum(tl.maximum(x_q, -448.0), 448.0)

    if NEEDS_MASK:
        tl.store(x_s_ptr + group_offsets, scale, mask=mask_1d)
        tl.store(x_q_ptr + offsets, x_q, mask=mask)
    else:
        tl.store(x_s_ptr + group_offsets, scale)
        tl.store(x_q_ptr + offsets, x_q)


def custom_kernel(data):
    x, x_q, x_s = data

    num_tokens, hidden_dim = x.shape
    num_groups = x_s.shape[1]
    group_size = hidden_dim // num_groups
    total_groups = num_tokens * num_groups

    # Cap BLOCK_GROUPS at 128. For a group_size of 128, this equates to 16,384 elements per block.
    # Distributed over 8 warps (256 threads), each thread holds 64 elements, sitting safely
    # below the A100 register bottleneck limits.
    BLOCK_GROUPS = 128

    # Scale down linearly for smaller dimensions to guarantee a deep dispatch grid.
    # Target ~512 blocks to allow ~4-5 wave parallelism across the 108 SMs.
    while BLOCK_GROUPS > 16 and (total_groups // BLOCK_GROUPS) < 512:
        BLOCK_GROUPS //= 2

    if BLOCK_GROUPS >= 128:
        num_warps = 8
    else:
        num_warps = 4

    grid = ((total_groups + BLOCK_GROUPS - 1) // BLOCK_GROUPS, )
    NEEDS_MASK = (total_groups % BLOCK_GROUPS) != 0

    fp8_quantize_kernel[grid](
        x, x_q, x_s,
        total_groups,
        GROUP_SIZE=group_size,
        BLOCK_GROUPS=BLOCK_GROUPS,
        NEEDS_MASK=NEEDS_MASK,
        num_warps=num_warps
    )

    return x_q, x_s
\end{Verbatim}
\end{pabox}
\end{document}